\documentclass[runningheads]{llncs}

\usepackage{eccv}

\usepackage{eccvabbrv}

\usepackage{graphicx}
\usepackage{booktabs}
\usepackage{amsmath}
\usepackage{amssymb}
\usepackage{bm}
\usepackage{booktabs}
\usepackage{multirow}
\usepackage{makecell}
\usepackage{xcolor}
\usepackage[accsupp]{axessibility}  

\usepackage[pagebackref,breaklinks,colorlinks,citecolor=eccvblue]{hyperref}

\usepackage{orcidlink}

\newcommand{\vect}[1]{\mathbf{#1}}
\newcommand{\vectsymb}[1]{\boldsymbol{#1}}
\newcommand{\mat}[1]{\mathbf{#1}}
\newcommand{\normal}{\mathcal{N}}
\newcommand{\expect}{\mathbb{E}}
\newcommand{\norm}[1]{\lVert #1 \rVert}

\begin{document}

\title{Diffusion-Refined VQA Annotations for Semi-Supervised Gaze Following} 

\titlerunning{GCDR-Gazefollowing}

\author{Qiaomu Miao \inst{1} \and
Alexandros Graikos \inst{1} \and
Jingwei Zhang \inst{1} \and Sounak Mondal \inst{1} \and \\
Minh Hoai \inst{2} \and Dimitris Samaras \inst{1} }

\authorrunning{Q. Miao et al.}

\institute{Stony Brook University, Stony Brook, USA \and The University of Adelaide, Adelaide, Australia
\email{\{qiamiao,agraikos,jingwezhang,somondal,minhhoai,samaras\}@cs.stonybrook.edu}}
\maketitle

\begin{abstract}
  Training gaze following models requires a large number of images with gaze target coordinates annotated by human annotators, which is a laborious and inherently ambiguous process. We propose the first semi-supervised method for gaze following by introducing two novel priors to the task. We obtain the first prior using a large pretrained Visual Question Answering (VQA) model, where we compute Grad-CAM heatmaps by `prompting' the VQA model with a gaze following question. These heatmaps can be noisy and not suited for use in training. The need to refine these noisy annotations leads us to incorporate a second prior. We utilize a diffusion model trained on limited human annotations and modify the reverse sampling process to refine the Grad-CAM heatmaps. By tuning the diffusion process we achieve a trade-off between the human annotation prior and the VQA heatmap prior, which retains the useful VQA prior information while exhibiting similar properties to the training data distribution. Our method outperforms simple pseudo-annotation generation baselines on the GazeFollow image dataset. More importantly, our pseudo-annotation strategy, applied to a widely used supervised gaze following model (VAT), reduces the annotation need by 50\%. 
  Our method also performs the best on the VideoAttentionTarget dataset. Code is available at \url{https://github.com/cvlab-stonybrook/GCDR-Gaze.git}
\end{abstract}

\def\mA{\mathcal{A}}
\def\mB{\mathcal{B}}
\def\mC{\mathcal{C}}
\def\mD{\mathcal{D}}
\def\mE{\mathcal{E}}
\def\mF{\mathcal{F}}
\def\mG{\mathcal{G}}
\def\mH{\mathcal{H}}
\def\mI{\mathcal{I}}
\def\mJ{\mathcal{J}}
\def\mK{\mathcal{K}}
\def\mL{\mathcal{L}}
\def\mM{\mathcal{M}}
\def\mN{\mathcal{N}}
\def\mO{\mathcal{O}}
\def\mP{\mathcal{P}}
\def\mQ{\mathcal{Q}}
\def\mR{\mathcal{R}}
\def\mS{\mathcal{S}}
\def\mT{\mathcal{T}}
\def\mU{\mathcal{U}}
\def\mV{\mathcal{V}}
\def\mW{\mathcal{W}}
\def\mX{\mathcal{X}}
\def\mY{\mathcal{Y}}
\def\mZ{\mathcal{Z}} 

\def\bbN{\mathbb{N}} 
\def\bbR{\mathbb{R}} 
\def\bbP{\mathbb{P}} 
\def\bbQ{\mathbb{Q}} 
\def\bbE{\mathbb{E}}

\def\1n{\mathbf{1}_n}
\def\0{\mathbf{0}}
\def\1{\mathbf{1}}

\def\A{{\bf A}}
\def\B{{\bf B}}
\def\C{{\bf C}}
\def\D{{\bf D}}
\def\E{{\bf E}}
\def\F{{\bf F}}
\def\G{{\bf G}}
\def\H{{\bf H}}
\def\I{{\bf I}}
\def\J{{\bf J}}
\def\K{{\bf K}}
\def\L{{\bf L}}
\def\M{{\bf M}}
\def\N{{\bf N}}
\def\O{{\bf O}}
\def\P{{\bf P}}
\def\Q{{\bf Q}}
\def\R{{\bf R}}
\def\S{{\bf S}}
\def\T{{\bf T}}
\def\U{{\bf U}}
\def\V{{\bf V}}
\def\W{{\bf W}}
\def\X{{\bf X}}
\def\Y{{\bf Y}}
\def\Z{{\bf Z}}

\def\a{{\bf a}}
\def\b{{\bf b}}
\def\c{{\bf c}}
\def\d{{\bf d}}
\def\e{{\bf e}}
\def\f{{\bf f}}
\def\g{{\bf g}}
\def\h{{\bf h}}
\def\i{{\bf i}}
\def\j{{\bf j}}
\def\k{{\bf k}}
\def\l{{\bf l}}
\def\m{{\bf m}}
\def\n{{\bf n}}
\def\o{{\bf o}}
\def\p{{\bf p}}
\def\q{{\bf q}}
\def\r{{\bf r}}
\def\s{{\bf s}}
\def\t{{\bf t}}
\def\u{{\bf u}}
\def\v{{\bf v}}
\def\w{{\bf w}}
\def\x{{\bf x}}
\def\y{{\bf y}}
\def\z{{\bf z}}

\def\balpha{\mbox{\boldmath{$\alpha$}}}
\def\bbeta{\mbox{\boldmath{$\beta$}}}
\def\bdelta{\mbox{\boldmath{$\delta$}}}
\def\bgamma{\mbox{\boldmath{$\gamma$}}}
\def\blambda{\mbox{\boldmath{$\lambda$}}}
\def\bsigma{\mbox{\boldmath{$\sigma$}}}
\def\btheta{\mbox{\boldmath{$\theta$}}}
\def\bomega{\mbox{\boldmath{$\omega$}}}
\def\bxi{\mbox{\boldmath{$\xi$}}}
\def\bnu{\mbox{\boldmath{$\nu$}}}                                  
\def\bphi{\mbox{\boldmath{$\phi$}}}
\def\bmu{\mbox{\boldmath{$\mu$}}}

\def\bDelta{\mbox{\boldmath{$\Delta$}}}
\def\bOmega{\mbox{\boldmath{$\Omega$}}}
\def\bPhi{\mbox{\boldmath{$\Phi$}}}
\def\bLambda{\mbox{\boldmath{$\Lambda$}}}
\def\bSigma{\mbox{\boldmath{$\Sigma$}}}
\def\bGamma{\mbox{\boldmath{$\Gamma$}}}
                                  
\newcommand{\myprob}[1]{\mathop{\mathbb{P}}_{#1}}

\newcommand{\myexp}[1]{\mathop{\mathbb{E}}_{#1}}

\newcommand{\mydelta}[1]{1_{#1}}

\newcommand{\myminimum}[1]{\mathop{\textrm{minimum}}_{#1}}
\newcommand{\mymaximum}[1]{\mathop{\textrm{maximum}}_{#1}}    
\newcommand{\mymin}[1]{\mathop{\textrm{minimize}}_{#1}}
\newcommand{\mymax}[1]{\mathop{\textrm{maximize}}_{#1}}
\newcommand{\mymins}[1]{\mathop{\textrm{min.}}_{#1}}
\newcommand{\mymaxs}[1]{\mathop{\textrm{max.}}_{#1}}  
\newcommand{\myargmin}[1]{\mathop{\textrm{argmin}}_{#1}} 
\newcommand{\myargmax}[1]{\mathop{\textrm{argmax}}_{#1}} 
\newcommand{\myst}{\textrm{s.t. }}

\newcommand{\denselist}{\itemsep -1pt}
\newcommand{\sparselist}{\itemsep 1pt}

\definecolor{pink}{rgb}{0.9,0.5,0.5}
\definecolor{purple}{rgb}{0.5, 0.4, 0.8}   
\definecolor{gray}{rgb}{0.3, 0.3, 0.3}
\definecolor{mygreen}{rgb}{0.2, 0.6, 0.2}

\newcommand{\cyan}[1]{\textcolor{cyan}{#1}}
\newcommand{\red}[1]{\textcolor{red}{#1}}  
\newcommand{\blue}[1]{\textcolor{blue}{#1}}
\newcommand{\magenta}[1]{\textcolor{magenta}{#1}}
\newcommand{\pink}[1]{\textcolor{pink}{#1}}
\newcommand{\green}[1]{\textcolor{green}{#1}} 
\newcommand{\gray}[1]{\textcolor{gray}{#1}}    
\newcommand{\mygreen}[1]{\textcolor{mygreen}{#1}}    
\newcommand{\purple}[1]{\textcolor{purple}{#1}}       

\definecolor{greena}{rgb}{0.4, 0.5, 0.1}
\newcommand{\greena}[1]{\textcolor{greena}{#1}}

\definecolor{bluea}{rgb}{0, 0.4, 0.6}
\newcommand{\bluea}[1]{\textcolor{bluea}{#1}}
\definecolor{reda}{rgb}{0.6, 0.2, 0.1}
\newcommand{\reda}[1]{\textcolor{reda}{#1}}

\def\changemargin#1#2{\list{}{\rightmargin#2\leftmargin#1}\item[]}
\let\endchangemargin=\endlist
                                               
\newcommand{\cm}[1]{}

\newcommand{\mhoai}[1]{{\color{magenta}\textbf{[MH: #1]}}}
\newcommand{\qiaomu}[1]{{\color{blue}\textbf{[QM: #1]}}}
\newcommand{\mtodo}[1]{{\color{red}$\blacksquare$\textbf{[TODO: #1]}}}
\newcommand{\myheading}[1]{\vspace{1ex}\noindent \textbf{#1}}
\newcommand{\htimesw}[2]{\mbox{$#1$$\times$$#2$}}


\newif\ifshowsolution
\showsolutiontrue

\ifshowsolution  
\newcommand{\Comment}[1]{\paragraph{\bf $\bigstar $ COMMENT:} {\sf #1} \bigskip}
\newcommand{\Solution}[2]{\paragraph{\bf $\bigstar $ SOLUTION:} {\sf #2} }
\newcommand{\Mistake}[2]{\paragraph{\bf $\blacksquare$ COMMON MISTAKE #1:} {\sf #2} \bigskip}
\else
\newcommand{\Solution}[2]{\vspace{#1}}
\fi

\newcommand{\truefalse}{
\begin{enumerate}
	\item True
	\item False
\end{enumerate}
}

\newcommand{\yesno}{
\begin{enumerate}
	\item Yes
	\item No
\end{enumerate}
}

\newcommand{\Sref}[1]{\cref{#1}}
\newcommand{\Eref}[1]{\cref{#1}}
\newcommand{\Fref}[1]{\cref{#1}}
\newcommand{\Tref}[1]{\cref{#1}}
\newcommand{\RN}[1]{%
  \textup{\uppercase\expandafter{\romannumeral#1}}%
}
\section{Introduction}
Human gaze behavior is a vital cue for understanding human cognitive processes \cite{yang2020invrf, yang2024topdown} and for applications such as human-machine interaction \cite{morimoto2005eye, drewes2010eye, admoni2017social}, social interaction analysis \cite{masse2017tracking, canigueral2019role}, and human intention interpretation \cite{sakita2004flexible, singh2020combining}. In contrast to eye-tracking glasses which are intrusive and mostly restricted to laboratory environments, the gaze following task \cite{NIPS2015_ec895663, chong2020detecting, fang2021dual} predicts a person's gaze target in an image in-the-wild, by predicting a target heatmap from an input scene image with the person's head bounding box. 

\begin{figure}[t]
\centering
\includegraphics[width=\linewidth]{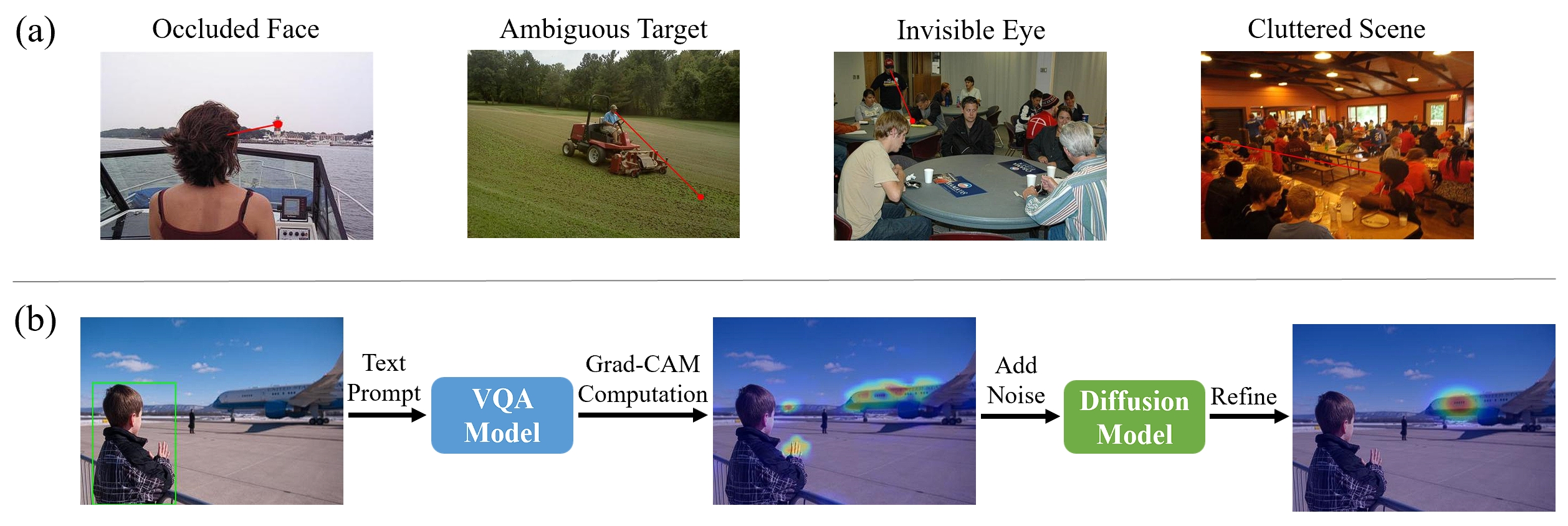}
\caption{(a) \textbf{Gaze following annotation challenges.} Annotating gaze is a laborious task with inherent ambiguities. (b) \textbf{Pseudo annotations for gaze following}. We generate pseudo annotations by first computing Grad-CAM heatmaps from a pre-trained VQA model, and then refining the noisy heatmaps with a diffusion model.} 
\label{fig:annotation_example}
\end{figure}

The effectiveness of gaze following methods depends on the quantity and quality of training data with annotated gaze targets. Annotating gaze targets is a laborious and ambiguous process. For an image and a subject in the image, an annotator must consider the subject's head orientation, gaze angle, and follow the gaze direction to determine where the point of gaze falls on an object or surface within the image. This can be very challenging, as the subject may be looking away from the camera, their eyes may be occluded, the image may be cluttered, and there may be multiple plausible targets, as shown in Fig.~\ref{fig:annotation_example}(a).

It is much more efficient to utilize large quantities of unlabeled data, along with a limited amount of labeled data, and leverage semi-supervised learning to improve model performance. Semi-supervised learning has been successful in tasks such as image recognition and semantic segmentation \cite{semisup_survey, meanteacher, kalluri2019universal, athiwaratkun2018there, hung2018adversarial}. 
In the context of gaze following, semi-supervised learning can be particularly useful when developing models for a specific scene, such as a psychology lab conducting social interaction studies or a supermarket studying customer gaze behaviors, by requiring only a small fraction of the collected data to be annotated.

This is the first semi-supervised gaze following method, to the best of our knowledge. Semi-supervised recognition and segmentation methods are typically not directly applicable to gaze following due to task differences. As gaze following models predict a target heatmap and require an intact scene image as input, methods that operate on the predicted multi-class distribution \cite{springenberg2015unsupervised, kalluri2019universal}, perform strong data augmentation (e.g.,CutOut\cite{sohn2020fixmatch,french2020semi}) cannot easily be adapted to gaze following. Additionally, for semi-supervised methods that use a teacher model to pseudo-annotate the unlabeled data \cite{lee2013pseudo, meanteacher, laine2017temporal}, it is unclear how to design a gaze following teacher model to generate good quality pseudo-annotations.

To generate high-quality pseudo-gaze labels, our semi-supervised method combines the power of pre-trained large vision-language (VL) models with an annotation refinement method. Due to their extensive training sets, VL models have naturally acquired a wealth of knowledge, including the knowledge for inferring human gaze targets. Specifically, we ``prompt'' a pre-trained Visual Question Answering (VQA) model~\cite{menon2022visual, radford2021learning, zhou2022learning} with an appropriate question to obtain the text description of the gaze target of a subject in the image. We then use Grad-CAM \cite{gradcan_paper, chefer2021generic}, to ground the VQA model's description to spatial locations, referred to as Grad-CAM heatmaps. We want to turn these heatmaps into pseudo-annotations for training gaze following models.

There are technical challenges in utilizing Grad-CAM heatmaps. First, Grad-CAM heatmaps do not precisely overlap with the gaze targets because they are dispersed and noisy, as shown in \Fref{fig:annotation_example}(b). Second, some Grad-CAM heatmaps do not provide useful prior knowledge and may highlight incorrect gaze target locations. Hence, the noisy Grad-CAM heatmaps should be refined into clean pseudo-annotations, with a trade-off between retaining the Grad-CAM heatmap information and the pre-trained prior on the annotations available.

We choose a diffusion model as the annotation prior of our refinement method. We are inspired by the recent use of diffusion models as inverse problem solvers and image editors that modify the reverse sampling process with a conditional input \cite{sdedit, song2022solving, diff_adversarialpurify}. The goal is to generate an output image that retains the semantics of the degraded input (here the Grad-CAM heatmap), while having similar properties (i.e., geometry and gaze context) to the training data distribution. 

We begin by training a diffusion model to capture the distribution of human-labeled annotations. The trained diffusion model then generates refined annotations by running the reverse sampling process, initialized from Grad-CAM heatmaps with an appropriate level of Gaussian noise. The added noise serves to smooth artifacts and noisy activations in the Grad-CAM heatmaps; the magnitude of the noise dictates how much of the information is preserved.

In summary, we propose the first semi-supervised gaze following method, in which we introduce two novel priors to the task: (1) Prior knowledge from pre-trained VQA models for initial annotation generation. (2) A novel diffusion-based annotation prior to refine the noisy VL annotations into reliable pseudo-labels.

Our method outperforms pseudo-annotation generation baselines on the GazeFollow image dataset \cite{NIPS2015_ec895663}. More importantly, our pseudo-annotation strategy, applied to the widely used VAT model \cite{chong2020detecting}, 
outperforms the fully supervised model trained with double the amount of annotations when only 5\% and 10\% labels are available. Our method also performs the best on the VideoAttentionTarget dataset \cite{chong2020detecting}, where we adapt a pre-trained gaze following model to new videos using only around 100 (2--23\%) annotated frames.

\section{Prior Work}

\myheading{Gaze Following} was first introduced in \cite{NIPS2015_ec895663} together with the GazeFollow dataset and a model composed of two separate pathways for encoding gaze orientation and scene saliency information, also adopted by later work \cite{lian2018believe, chong2018connecting, chong2020detecting, fang2021dual}.
Chong \etal \cite{chong2018connecting} considered out-of-frame targets and also extended the task to videos \cite{chong2020detecting}, proposing the VideoAttentionTarget dataset and the VAT model. Later work has improved gaze following models by leveraging monocular depth estimations \cite{fang2021dual, jin2022depth, tonini2022multimodal, hu2022we} and human poses \cite{gupta2022modular, Bao_2022_CVPR}. Tu \etal \cite{tu2022end} proposed a transformer model for gaze following, while \cite{jin2021multi,miao2023patch} added training losses for numerical coordinate regression and patch-level gaze distribution prediction. Recent works investigate 3D gaze following \cite{gf3d_cvpr23, gf3d_cvprw23}, gaze following for children \cite{tafasca2023childplay}, and object-aware gaze target prediction \cite{tonini_iccv23}. All these methods are fully-supervised. A concurrent work investigated semi-supervised gaze following using a saliency prediction model \cite{peng2024semi}. This approach does not leverage the strong priors from pretrained VL models, and its performance falls significantly behind our method.

\myheading{Semi-supervised Learning} methods mostly adopt a teacher-student framework, which can be categorized into self-training \cite{lee2013pseudo, chen2022debiased, grandvalet2004semi, sohn2020fixmatch} and consistency regularization \cite{laine2017temporal, meanteacher, french2020semi, ho2024diffusion}. Self-training methods generate pseudo annotations from teacher models trained with labeled data, while consistency regularization applies a consistency loss between the student and teacher model outputs, updating the teacher model gradually during training. Due to task differences, most semi-supervised recognition or segmentation methods are not directly applicable to gaze following \cite{sohn2020fixmatch, paige2017learning, french2020semi, kalluri2019universal, li2019disentangled}. However, some general methods without task-specific operations \cite{laine2017temporal, meanteacher} are applicable with appropriate modifications (see Supplementary). In our method, we adopt a self-training pipeline using the VQA prior and diffusion-based annotation prior, and show that we can get further improvements by using applicable consistency regularization methods (e.g. Mean Teacher \cite{meanteacher}) to enhance the diffusion model training. 

\myheading{Vision-Language Models} have prospered after the development of large language models (LLMs) \cite{touvron2023llama, devlin2018bert, chowdhery2023palm, lyu2023attention, lyu2022study}, and have succeeded in tasks such as visual grounding, visual question answering (VQA), and image captioning \cite{wang2022ofa, kamath2021mdetr, singh2022flava, li2022blip, liu2023visual}. They also do well in low-shot generalization tasks by ``prompting''~\cite{pratt2022does, menon2022visual, gazeformer}, where task instructions are given to a pretrained model to generate outputs useful for other tasks. Grad-CAM~\cite{selvaraju2017grad} visualizations \textit{localize} the linguistic input effect on the visual input \cite{li2021align, zeng2021multi} thus achieving a degree of interpretability.

\myheading{Diffusion Models}. Denoising diffusion probabilistic models (DDPM) \cite{sohl2015deep, ho2020denoising} give state-of-the-art results on image synthesis \cite{dhariwal2021diffusion, rombach2022highresolution, graikos2022diffusion, diffusion_seg_refine,  Graikos_2024_CVPR}.
They are also used in inverse problems \cite{song2022solving}, image editing \cite{sdedit}, and adversarial purification \cite{diff_adversarialpurify}. These works treat images with added noise as intermediate steps of the reverse process, showing that the noise magnitude affects the information retained from the input. Recent work adopts diffusion models in semi-supervised learning. \cite{ho2024diffusion} trains a diffusion model for 3D object detection, whereas \cite{you2024diffusion} trains a diffusion model for image generation with images pseudo-labeled by existing semi-supervised models. In contrast, we use diffusion model as annotation prior, to refine the initial noisy annotations into high-quality pseudo annotations.

\section{Proposed Method}

\subsection{Overview}

\begin{figure}[t]
\centering
\includegraphics[width=\textwidth]{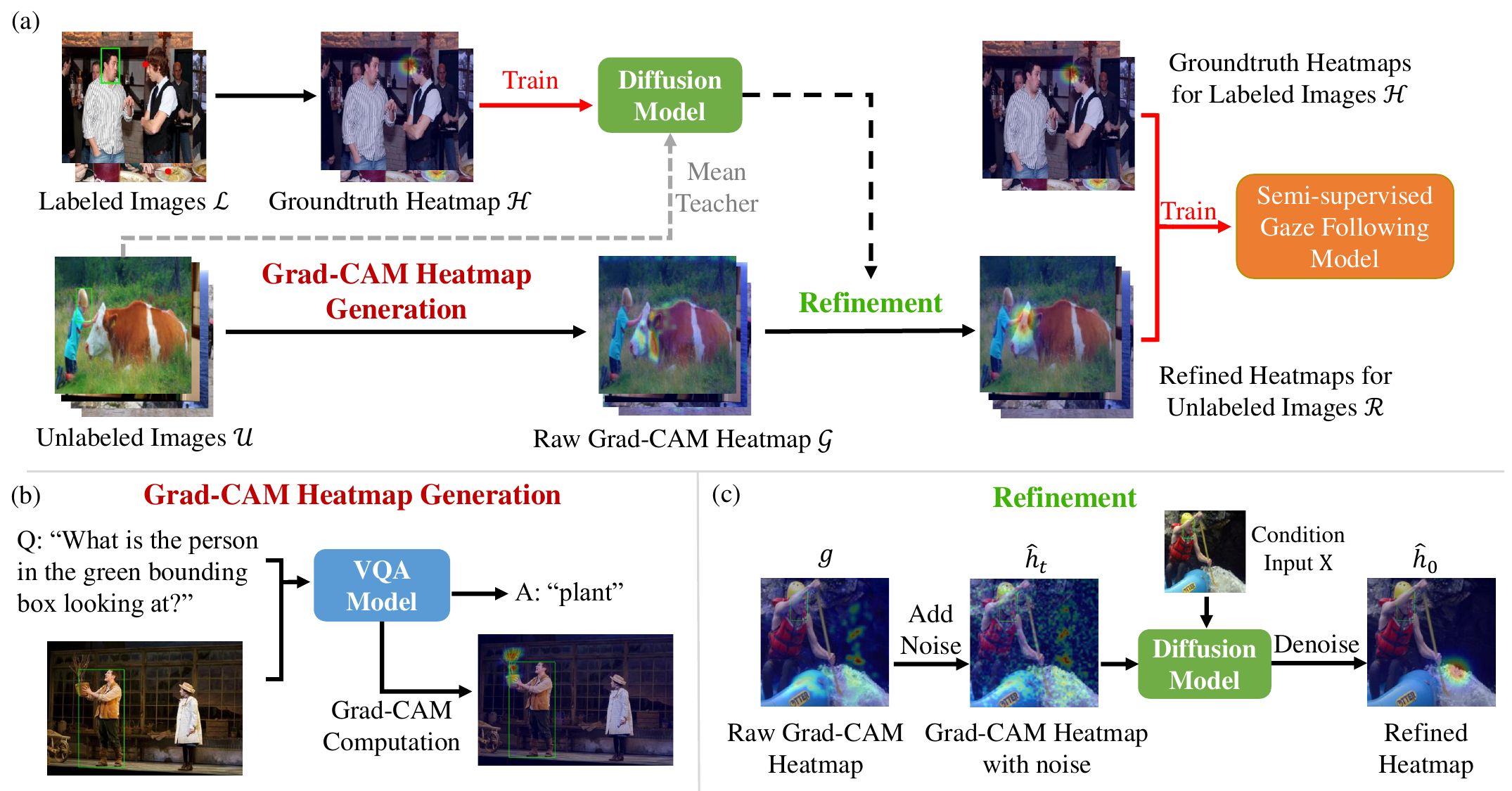}
\caption{\textbf{(a) Overall pipeline}. We compute Grad-CAM heatmaps for unlabeled images and train the diffusion model with a small human-labeled set (or with unlabeled images using Mean Teacher). The diffusion model refines the Grad-CAM heatmaps into pseudo-annotations. Both the pseudo-annotations and the human-labeled set are used to train a gaze following model. \textbf{(b) Grad-CAM heatmap generation.} Given an image with an overlaid person bounding box, we ``prompt" a pretrained VQA model with a gaze question and compute the Grad-CAM heatmap from the answer. \textbf{(c) Grad-CAM refinement.} We perturb the Grad-CAM heatmaps with Gaussian noise and pass through the reverse diffusion process to generate the refined pseudo-annotations.
}
\label{fig:framework}
\end{figure}

Our approach is illustrated in \Fref{fig:framework}a. The goal is to leverage the knowledge priors from a VQA model to improve gaze following with minimal manual annotation. To achieve this, we utilize Grad-CAM heatmaps of a VQA model that is prompted with gaze-related questions. 

However, the Grad-CAM heatmaps need refinement to be useful for training supervision. For this, we train a diffusion model on the groundtruth heatmaps from labeled data to learn to generate heatmaps from the human-label distribution. We use the model to refine the ``noisy'' Grad-CAM into high-quality pseudo annotations. We train existing gaze following models on both groundtruth heatmaps from labeled data and refined heatmaps from unlabeled data.

Suppose we only have a small number of images that are labeled with gaze target locations ${\mathcal{L}=\{(\bm{X}_{i}, \bm{p}_i)\}_{i=1}^{N_l}}$ and a larger number of unlabeled images ${\mathcal{U}=\{\bm{X}_{j}\}_{j=1}^{N_u}}$. $\bm{X}_i$ indicates an input triplet to the gaze following models, which includes a scene image $\bm{I}_{i} \in R^{3 \times H \times W}$, the cropped head image $\bm{I}_h^i \in R^{3 \times H \times W}$, and the head location binary mask $\bm{M}_h^i \in R^{H \times W}$. We represent the annotated gaze coordinate of the "gazing" person as $\bm{p}_i = [p_i^x, p_i^y]$. 
For each $\bm{p}_i$, a ground truth heatmap $\vect{h}^i \in R^{H' \times W'}$ is generated by placing a Gaussian of fixed variance centered at ${\bm{p}_i}$, as is done in previous gaze following frameworks \cite{chong2020detecting, fang2021dual,lian2018believe}. 

\subsection{Grad-CAM heatmap extraction} \label{gradcam_gen}

We use OFA \cite{wang2022ofa}, a transformer-based large pretrained VL model as the VQA model. \Fref{fig:framework}b illustrates the procedure for Grad-CAM heatmap generation. Given an image $\bm{I}$ and the head bounding box of a person $\bm{l}_h$, we use Mask R-CNN \cite{he2017mask} for person detection and find the bounding box that maximally overlaps with $\bm{l}_h$. We overlay this bounding box on top of the input image and give it as input to the VQA model along with a ``prompt'' in the form of a question: {``\it What is the person in the green bounding box looking at?"}. 

OFA has an encoder and a decoder. It generates a sequence of words as the answer. We compute Grad-CAM heatmaps $\mathcal{G}=\{\bm{g}_{j}\}_{j=1}^{N_u}$ on the decoder cross-attention weights between the input query and the image patch tokens based on \cite{chefer2021generic} after selecting the noun from the answer. Details are in the supplementary.

In our experiments, we found that Grad-CAM heatmaps are beneficial for gaze following. Naively using these heatmaps as additional input to the current gaze following model results in a significant performance boost. However, in real applications, computing Grad-CAM heatmaps at test-time can be impractical due to the memory and time costs of accessing large VL models.
Instead we use  Grad-CAM heatmaps offline as pseudo-annotations for training a gaze following model. As  discussed, using noisy Grad-CAMs directly as pseudo labels can hurt final model performance. Thus, we propose to utilize a diffusion model to refine the initial Grad-CAM heatmaps into more suitable pseudo labels for training.

\subsection{Diffusion Model Training}

As there is no available diffusion model for the gaze following task, we first train a diffusion model on the labeled data to generate gaze heatmaps $\vect{h} \sim p(\vect{h|\bm{X}})$, following the general training procedure of DDPM \cite{ho2020denoising}. For ease of understanding, we provide  a synopsis of Gaussian diffusion models. 

\begin{figure}[t]
\centering
\includegraphics[width=0.95\linewidth]{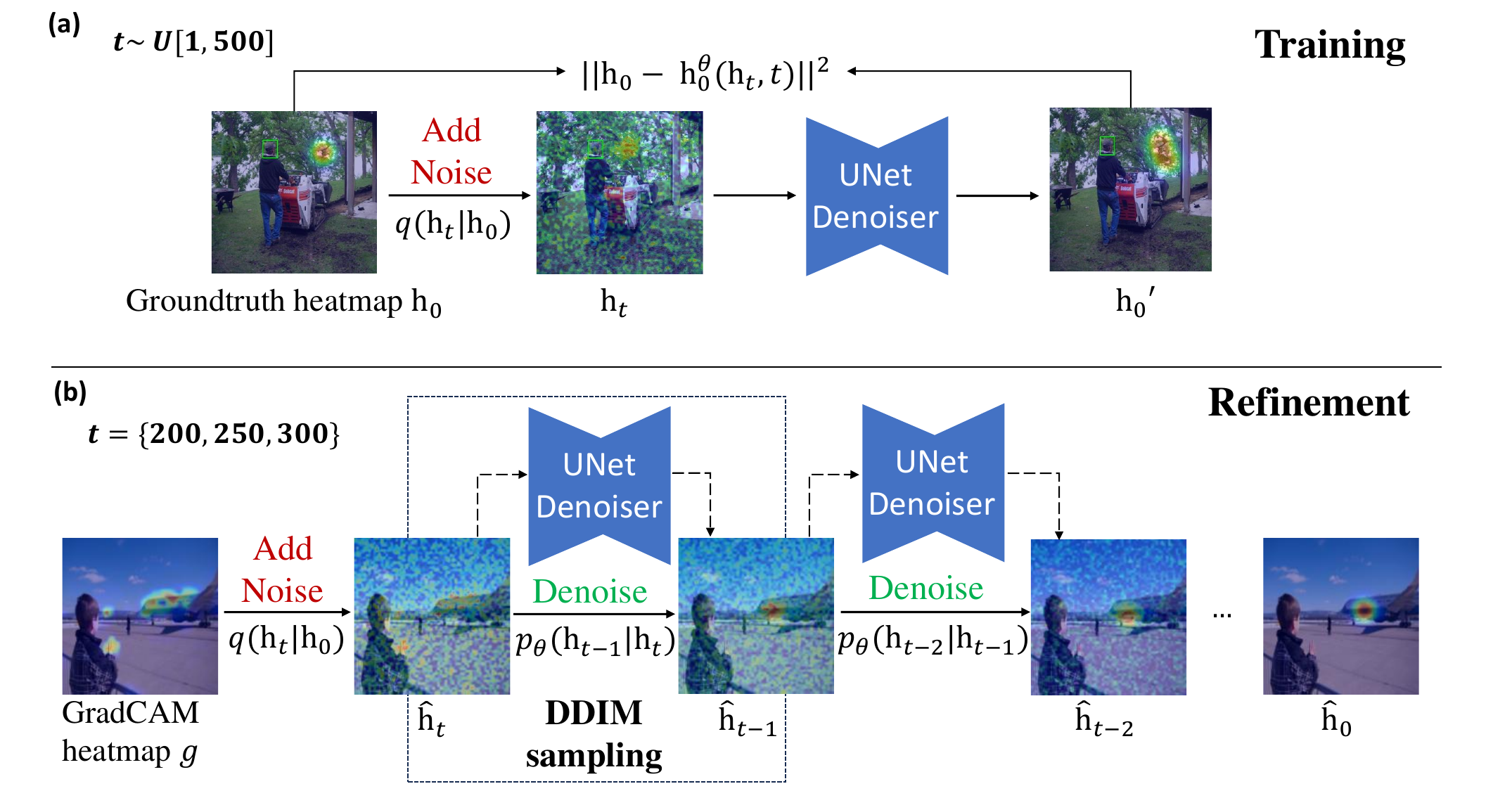}
\caption{\textbf{Diffusion model training and refinement}. (a) The diffusion model is trained on supervised data with noise added to the ground truth heatmap at random time steps. (b) During refinement, we add noise at a specific time step to the Grad-CAM heatmap. We treat this heatmap as an intermediate step input during the reverse process. Heatmaps are overlayed on the original images for illustration purposes. The conditional feature extraction for the diffusion model is omitted for simplicity.}
\label{fig:diffusion_model}
\end{figure}

The diffusion model consists of a forward process that gradually corrupts the input by adding Gaussian noise, and a reverse process that iteratively denoises the noisy input. The forward process is defined by a noise schedule $\alpha_t$ as:
\begin{align}
    q(\vect{h}_t|\vect{h}_{t-1}) &:= \normal(\vect{h}_t; \sqrt{\alpha_t}\vect{h}_{t-1}, (1-\alpha_t)\mat{I}) \\
    q(\vect{h}_{1:T}|\vect{h}_0) &:= \prod_{t=1}^T q(\vect{h}_{t-1}|\vect{h}_t).
    \label{eq:fw_process}
\end{align}
where $\vect{h}_0$ is the ground truth heatmap $\vect{h}$, $\vect{h}_{1},\dots,\vect{h}_{T-1}$ are intermediate latent variables that represent noisy versions of $\vect{h}_0$, and $\vect{h}_T$ is the terminal state which corresponds to a unit Gaussian distribution: $\vect{h}_T \sim \normal (\vect{0},\mat{I})$.

The reverse process uses Gaussian transitions with fixed covariance $\sigma_t$ and gradually denoises the data, starting from $\vect{h}_T$. Since $q(\vect{h}_{t-1}|\vect{h}_{t})$ is intractable, it is approximated with a neural network:
\begin{equation}
    p_{\theta}(\vect{h}_{t-1}\mid\vect{h}_t) := \normal(\vect{h}_{t-1}; \vectsymb{\mu}_{\theta}(\vect{h}_t, t), \sigma_t \mat{I}).
    \label{eq:reverse_process}
\end{equation}
A useful property of the forward process is that it allows sampling of any intermediate $\vect{h}_t$ given $\vect{h}_0$, and $\bar{\alpha}_t = \prod_{i=1}^t \alpha_i$:
\begin{equation}
    q(\vect{h}_t|\vect{h}_0) = \normal(\vect{h}_t; \sqrt{\bar{\alpha}_t}\vect{h}_0, (1-\bar{\alpha}_t)\mat{I})
    \label{eq:xt_sampling}
\end{equation}
which is utilized in learning the reverse process. For that, we adopt the $\vect{x}_0$-parameterization as in \cite{austin2021structured, diffusion_seg_refine} and directly predict the final~$\vect{h}_0$, leading to the following minimization objective, where $\vect{h}_{0}^{\theta}$ is parameterized by a U-Net~\cite{ronneberger2015unet} :
\begin{equation}
    \expect_{\vect{h}_0, t} \left[\norm{\vect{h}_0 - \vect{h}_{0}^{\theta}(\vect{h}_t, t)}^2 \right],
    \label{eq:ddpm_obj}
\end{equation}

\Fref{fig:diffusion_model}(a) shows the training procedure. At each iteration, we sample a random time step $t$ from $U[1, T]$ to generate a noisy input $\vect{h}_t$ using \Eref{eq:xt_sampling}. We feed $\vect{h}_t$ to the diffusion model for single-step denoising, and optimize with the loss of \Eref{eq:ddpm_obj}. We condition the diffusion model with the gaze feature $\vect{c}$ extracted from the input triplet $\bm{X}$ (input image, head crop, and head location mask) to output a gaze heatmap grounded on the gazing person; the parametrization of the U-Net denoiser becomes $ \vect{h}_{0}^{\theta}(\vect{h}_t, t, \vect{c})$. The feature extractor follows the structure of the VAT model \cite{chong2020detecting}, which takes the input triplet $\bm{X}$ and outputs the extracted features concatenated from two pathways for encoding the scene and gaze features. Details of the VAT structure are shown in Supplementary. Following~\cite{giannone2022few}, we add the extracted features as conditional features to each layer in the U-Net denoiser with the necessary transposed convolution layers to match the feature size of U-Net. The diffusion model can also be further improved by training with unlabeled data using Mean Teacher (MT) \cite{meanteacher}, a general-purpose semi-supervised learning method. As a consistency regularization method, MT enforces consistencies between the outputs from the teacher and student models fed with differently perturbed input. The teacher model has the same structure as the student model and is updated with an Exponential Moving Average (EMA) of the student model weights during training. In our case, we treat the diffusion model as the student model and apply different color jittering and head bounding box jittering to introduce different perturbations to the teacher and student model input. During training, the diffusion model is trained with the loss on labeled data and an additional consistency loss between the output heatmaps from the teacher and student diffusion models on all data.  We demonstrate in our experiments that this further improves the gaze following results.

\subsection{Heatmap Refinement using Diffusion Models}

After training on labeled images $\mathcal{L}$, the learned diffusion process models an annotation prior over ground truth annotations $p(\vect{h|\bm{X}})$.
During the semi-supervised training process, the trained diffusion model is applied to the unsupervised data $\mathcal{U}$ to produce the refined pseudo annotations.

The refinement process is shown in \Fref{fig:diffusion_model}(b).  Given a noisy Grad-CAM heatmap $\bm{g}$ of an unsupervised input sample $\bm{X}$, we first add an appropriate amount of noise using the forward diffusion process $\hat{\vect{h}}_t = \sqrt{\bar{\alpha}_t}\bm{g} + \sqrt{1-\bar{\alpha}_t}\vectsymb{\epsilon}$ according to \Eref{eq:xt_sampling}. The added noise smoothes out the noise artifacts of the Grad-CAM heatmaps, while preserving the information of the highlighted regions. Then, by iteratively applying the denoising steps in \Eref{eq:reverse_process}, we get samples of the final heatmap $\hat{\vect{h}}_0$ that are geometrically and contextually similar to the human-labeled heatmap $\vect{h}_0$ while retaining gaze target information from the noisy Grad-CAM input. During the denoising process, we used DDIM \cite{ddim} to sample the next step input using the predicted $\vect{h}_0$ from the U-Net. 

The amount of information retained depends on the choice of $t$ (\ie, the magnitude of added noise), as proven in \cite{sdedit, diff_adversarialpurify}. 
A larger magnitude of noise corrupts the Grad-CAM heatmaps more, decreasing the similarity between the diffusion output and the Grad-CAM priors. Similar to \cite{sdedit} which achieves a good trade-off between \textit{realism} and \textit{faithfulness} in image editing with an intermediate level of noise, we also found that initializing from an intermediate timestep in the diffusion process achieved the best trade-off between the Grad-CAM information and the model's learned distribution $p(\vect{h|\bm{X}})$. With this formulation, the diffusion model can incorporate the high-quality Grad-CAM heatmaps, while ignoring the Grad-CAM priors when they are highly noisy or highlight unlikely gaze target locations. We provide analyses of the added noise magnitude in \Sref{sec:test_noise}.

\subsection{Semi-supervised Training}

In semi-supervised training, we use the refined heatmaps $\mathcal{R}$ as pseudo labels for the unsupervised data $\mathcal{U}$. We train the gaze following model with the set of ground-truth heatmaps $\mathcal{H}$ from the labeled data $\mathcal{L}$ and the refined heatmaps $\mathcal{R}$ from the unlabeled data $\mathcal{U}$. We use the Mean Squared Error (MSE) loss between the predicted heatmap $\bm{h'}$ and the target label $\h$. $\h$ is either a human annotation (labeled data) or a pseudo annotation (unlabeled data). We minimize:
\begin{equation}
    \mathcal{L} = \frac{1}{|\mathcal{H} \cup  \mathcal{R}|} \sum_{\h \in \mathcal{H} \cup \mathcal{R} }^{}\mathcal{L}_{mse}(\bm{h'}, \bm{h}).
\end{equation}

\section{Experiments}
\subsection{Setting and Implementation Details}
\myheading{Datasets.} \textbf{GazeFollow} \cite{NIPS2015_ec895663}, the largest real image dataset, contains the annotated gaze targets of 130,339 people in 122,143 images. 4782 people are used for testing, and the rest are used in training. For the test set, there are 10 annotations per person-image pair to account for the ambiguity of the gaze target, whereas the training set contains a single annotation for each pair. \textbf{VideoAttentionTarget} \cite{chong2020detecting} is a video dataset of 50 different shows collected from YouTube. Each person-image pair in both training and test set has only one annotation.

\myheading{Evaluation metrics.} We employ  evaluation metrics suggested by previous work~\cite{NIPS2015_ec895663, chong2020detecting, lian2018believe}. The distance metric (\textbf{Dist.}) refers to the normalized $L_2$ distance between the predicted gaze point (point with the highest heatmap response) and the ground truth location. In GazeFollow reports  both average (\textbf{AvgDist}) and minimum distances (\textbf{MinDist}). Area Under Curve (\textbf{AUC}) evaluates the concordance of predicted heatmaps with ground truth \cite{chong2020detecting}. For GazeFollow, ground truth is the 10 annotations for each test set image, whereas in VideoAttentionTarget it is a thresholded Gaussian centered at the single given annotation. 

\myheading{Implementation Details.}
We used a diffusion model with a linear noise schedule and $T{=}500$ steps. The heatmap size was $64 {\times} 64$. We used a batch size of 80 and a learning rate of $2.5 \times 10^{-4}$ in the semi-supervised learning experiments. For training the diffusion model on labeled data, we used a learning rate of $5 \times 10^{-5}$. When refining Grad-CAM heatmaps, we used DDIM~\cite{ddim} for inference.

\subsection{Semi-supervised Training Results} \label{sec:semi_results}

We build our gaze following experiments around the popular, publicly available VAT model \cite{chong2020detecting}. To showcase the impact of our pseudo annotation generation method, we did not use additional modalities, processing, or supervisions, such as depth input, pose estimations, and object-level annotations~\cite{fang2021dual, Bao_2022_CVPR, gupta2022modular, tonini_iccv23} in all experiments for the baseline methods and the diffusion model.

Following the standard experimental settings in semi-supervised learning \cite{french2020semi, lee2013pseudo, meanteacher}, we considered different amounts of annotations from the GazeFollow training set, namely 5\%, 10\%, and 20\%, and treated the remaining data as unlabeled. The trained student model is evaluated on the test set in all cases.

We used VAT as the student model and compared our method of generating pseudo annotations with several alternatives\footnote{Implementation details of the baselines are provided in the supplementary material.}:  
1) {\it Semi-VAT}: the pseudo-annotations for the unlabeled data were generated by the VAT model trained with labeled images. 2) {\it Semi-VAT-GC}: We created VAT-GC by adding Grad-CAM heatmaps as conditional inputs to the VAT model and modifying the VAT model accordingly. VAT-GC was trained on the labeled data to generate pseudo annotations for the unlabeled data. 3) {\it VAT-MT}: We use the Mean Teacher (MT) method \cite{meanteacher} to train the VAT model. The first two baselines are self-training methods, while the 3rd belongs to the consistency regularization category. 

On the other hand, we also build two versions of our method: 1) {\it GradCAM-Diffusion-Refinement (GCDR)}: our method of using a diffusion model trained with labeled data to refine the `noisy' Grad-CAM heatmaps and use them as pseudo-annotations. 2) {\it GCDR-MT}: we used the Mean-Teacher method to train the diffusion model with both labeled and unlabeled data, and this enhanced diffusion model was used to refine the Grad-CAM heatmaps.

\begin{table}[t]
\centering
\caption{\textbf{Results of semi-supervised training methods with different ratios of labeled data on GazeFollow dataset}. In the top and bottom rows, we show the performances of VAT trained with supervised data only or with full training data, to show the potential lower and upper limits. Best numbers are marked as bold. Our methods outperform supervised VAT trained with double the amount of annotations.}
\footnotesize
\begin{tabular}{l|ccccccccc}
\Xhline{2\arrayrulewidth}
 Method  & \multicolumn{3}{c|}{5\% labels}                                            & \multicolumn{3}{c|}{10\% labels}                                            & \multicolumn{3}{c}{20\% labels}                                           \\ \cline{2-10} 
                  & \multicolumn{2}{c}{Dist. $\downarrow$}  &  \multicolumn{1}{c|}{\multirow{2}{*}{AUC $\uparrow$}}           & \multicolumn{2}{c}{Dist $\downarrow$}                 & \multicolumn{1}{c|}{\multirow{2}{*}{AUC $\uparrow$}}          & \multicolumn{2}{c}{Dist. $\downarrow$}                  & \multirow{2}{*}{AUC $\uparrow$} \\
                                           & \multicolumn{1}{l}{Avg.} & \multicolumn{1}{l}{Min.} &    \multicolumn{1}{c|}{}       & \multicolumn{1}{l}{Avg.} & \multicolumn{1}{l}{Min.} & \multicolumn{1}{c|}{}  & \multicolumn{1}{l}{Avg.} & \multicolumn{1}{l}{Min.} & \\ \Xhline{2\arrayrulewidth}
VAT (Supervised)    & 0.230                    & 0.161 & \multicolumn{1}{c|}{0.835}         &        0.202                    & 0.133         & \multicolumn{1}{c|}{0.869}                &           0.182                    & 0.116            & 0.892                                    \\
\Xhline{\arrayrulewidth}                

Semi-VAT   & 0.222               & 0.152  & \multicolumn{1}{c|}{0.846}         & 0.195                    & 0.128           & \multicolumn{1}{c|}{0.875}                 & 0.176                    & 0.110          & 0.896                                            \\
Semi-VAT-GC  & 0.217                &  0.149  & \multicolumn{1}{c|}{0.846}             & 0.195                    & 0.127       & \multicolumn{1}{c|}{0.879}                   & 0.178                    & 0.112       & 0.895                                            \\
VAT-MT  &              0.219                &      0.150 &   \multicolumn{1}{c|}{0.850}        & 0.189                    & 0.122           & \multicolumn{1}{c|}{0.882}                     & 0.174                    & 0.108     & 0.898                             \\  
GCDR (Ours)  &  0.201    &  0.135 &  \multicolumn{1}{c|}{0.863} &    0.179    &    0.115   &  \multicolumn{1}{c|}{0.886}  &   0.166    &  0.103    & 0.902                \\ 
GCDR-MT (Ours)  &   \textbf{0.194}    &    \textbf{0.128} & \multicolumn{1}{c|}{\textbf{0.870}}  &    \textbf{0.172}    &   \textbf{0.108} &   \multicolumn{1}{c|}{\textbf{0.892}}     &  \textbf{0.162}     &   \textbf{0.098}   & \textbf{0.904}                       \\ 
\Xhline{\arrayrulewidth}  
VAT(100\% labels) \cite{chong2020detecting}   & & & & 0.137 & 0.077 & 0.921 & & & \\
\Xhline{2\arrayrulewidth}
\end{tabular}
\label{tab:semi_results}
\end{table}

Semi-supervised training results  are  in \Tref{tab:semi_results}. Our two proposed methods consistently outperform the three semi-supervised baselines in all three annotation scenarios. When training with 5\% and 10\% annotations, the performance of \textit{GCDR-MT} surpasses the supervised VAT trained with double the amount of annotations. Results of training with 50\%  are in the supplementary.

\begin{figure}[t]
\centering
\setlength\tabcolsep{0.5pt}
\def\arraystretch{0.50}
\begin{tabular}{ccccc}
 VAT &
 VAT-GC &
 Raw Grad-CAM &
 GCDR &
 Groundtruth\\
\includegraphics[width=0.19\linewidth, height=0.65in]{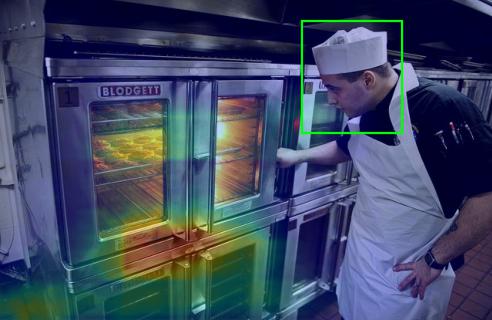}&
 \includegraphics[width=0.19\linewidth, height=0.65in]{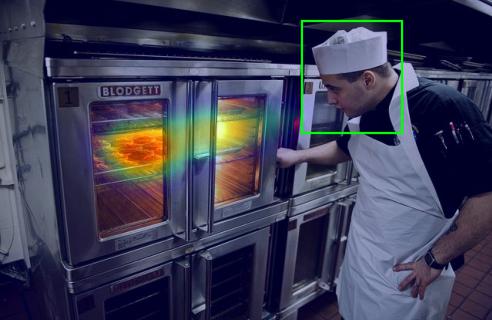} &
 \includegraphics[width=0.19\linewidth, height=0.65in]{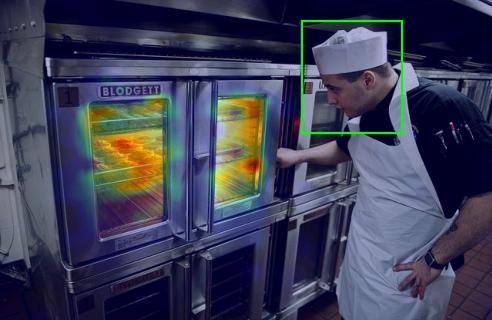} &
 \includegraphics[width=0.19\linewidth, height=0.65in]{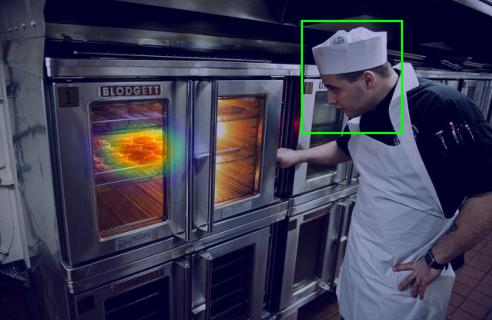} &
 \includegraphics[width=0.19\linewidth, height=0.65in]{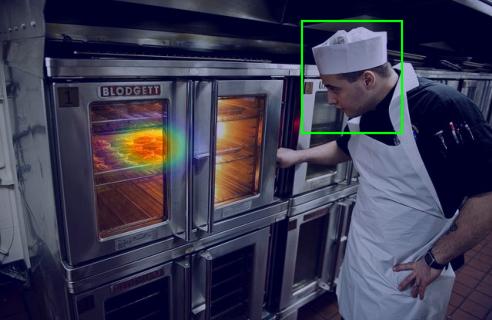}  \\
  \includegraphics[width=0.19\linewidth, height=0.65in]{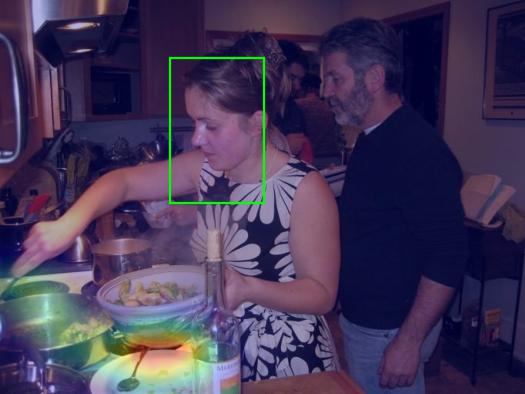}&
 \includegraphics[width=0.19\linewidth, height=0.65in]{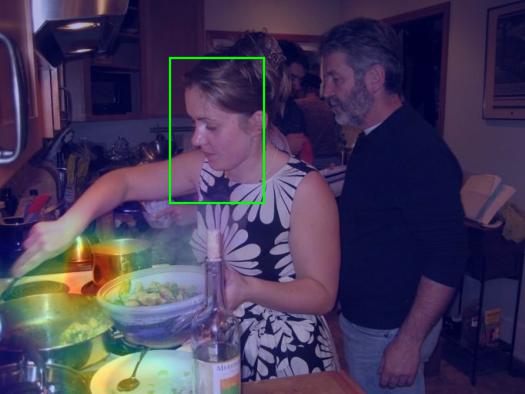} &
 \includegraphics[width=0.19\linewidth, height=0.65in]{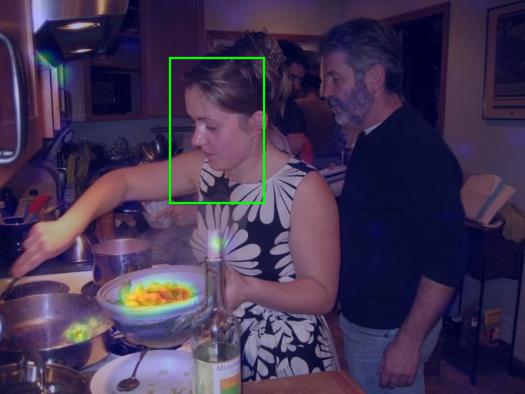} &
 \includegraphics[width=0.19\linewidth, height=0.65in]{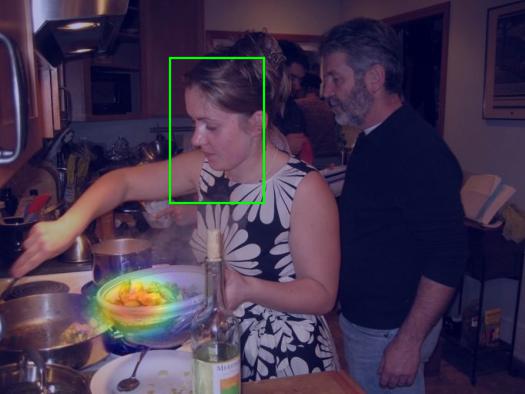} &
 \includegraphics[width=0.19\linewidth, height=0.65in]{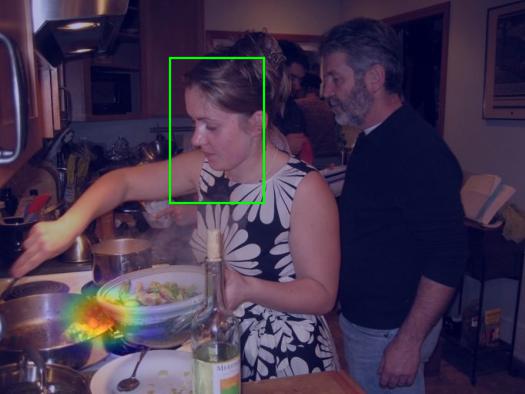} \\
\includegraphics[width=0.19\linewidth, height=0.65in]{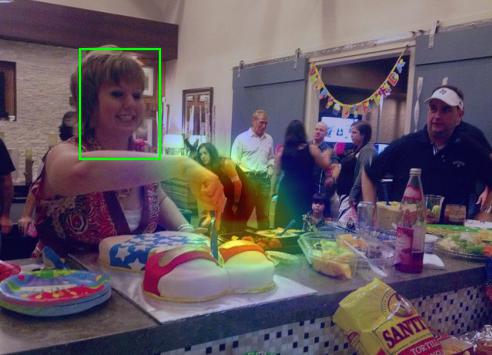}&
 \includegraphics[width=0.19\linewidth, height=0.65in]{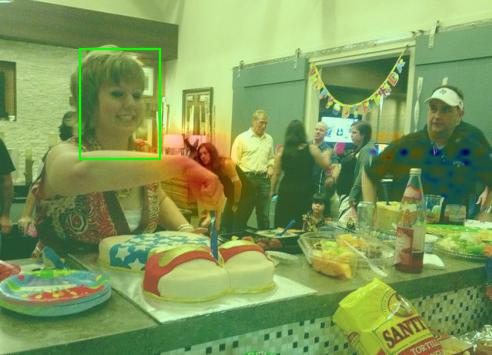} &
 \includegraphics[width=0.19\linewidth, height=0.65in]{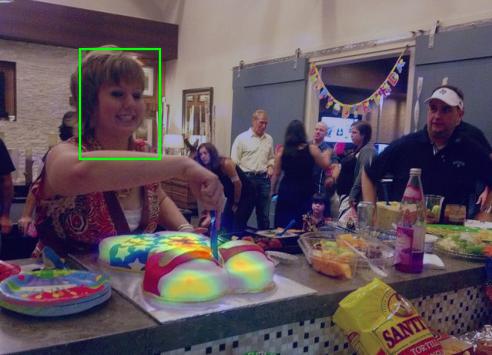} &
 \includegraphics[width=0.19\linewidth, height=0.65in]{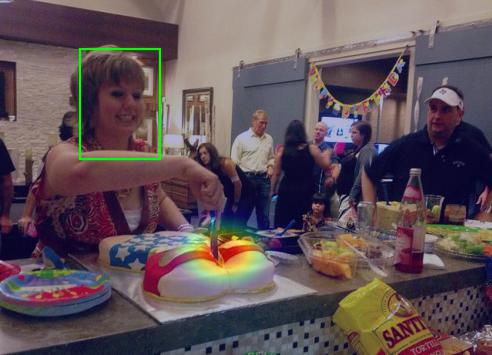} &
 \includegraphics[width=0.19\linewidth, height=0.65in]{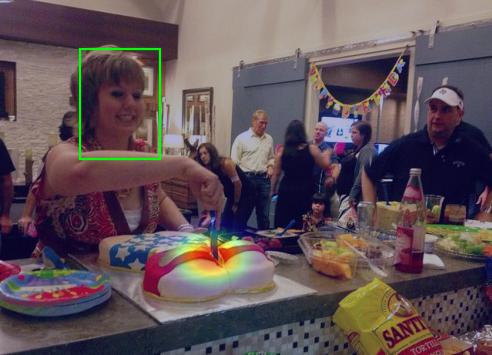} \\
 \includegraphics[width=0.19\linewidth, height=0.65in]{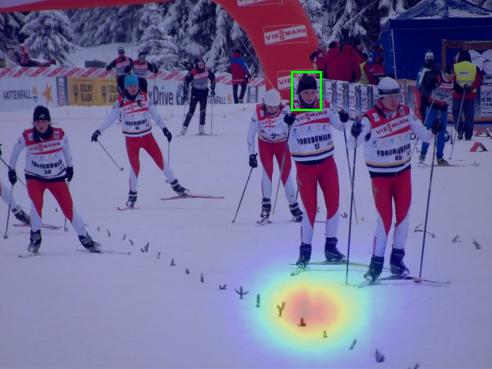}&
 \includegraphics[width=0.19\linewidth, height=0.65in]{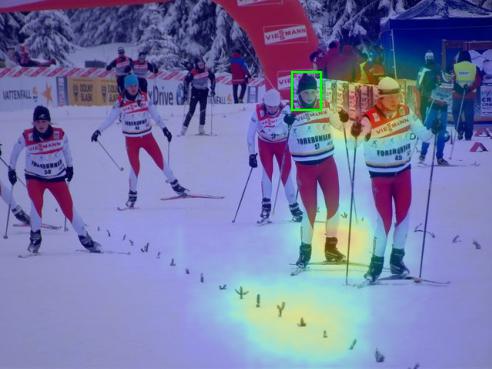} &
 \includegraphics[width=0.19\linewidth, height=0.65in]{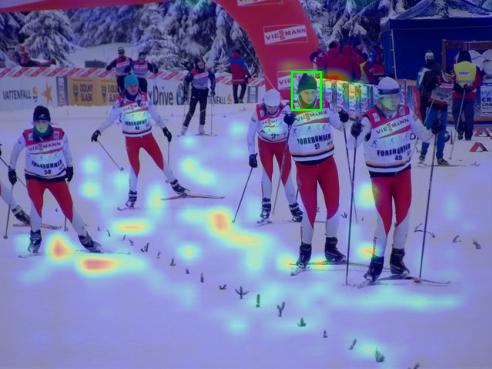} &
 \includegraphics[width=0.19\linewidth, height=0.65in]{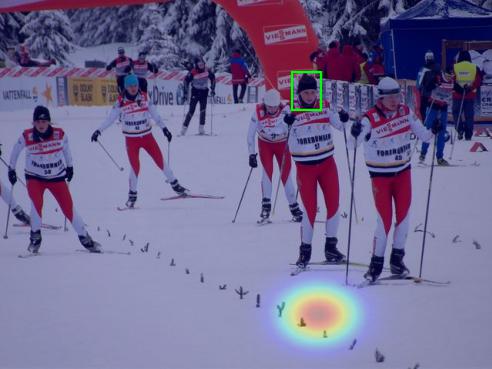} &
 \includegraphics[width=0.19\linewidth, height=0.65in]{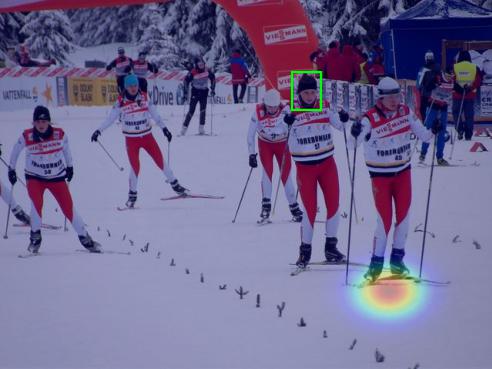} 
\end{tabular}
\caption{\textbf{Visualizations of pseudo heatmaps generated by different teachers}. Our method generates the cleanest pseudo annotations while retaining the Grad-CAM heatmaps priors (Rows 1--3). When the initial Grad-CAM heatmap responds strongly to unlikely locations or is completely noisy, our method can also ignore it (Row 4).}
\label{fig:qualitative_vis}
\end{figure}

Our methods show more prominent improvements in Dist. compared to AUC. The AUC on GazeFollow evaluates the concordance of the predicted heatmap with the 10 annotations on each test image, while Dist. evaluates the $L_2$ distance between the predicted gaze point and the annotations. Therefore, the results mean that we did better in predicting the probable target location than predicting the exact shape of the heatmap. The refined heatmaps from the diffusion model closely resemble the spatial distribution of a human-labeled annotation, which is a single Gaussian (\Fref{fig:qualitative_vis}). In contrast, the baseline methods predict the target location less accurately, while outputting less certain and larger heatmaps which tend to overlap more with the group-level annotations, thus favoring the AUC (see supplementary material).

\Fref{fig:qualitative_vis} visualizes the pseudo annotations predicted by the teacher models. In the top three rows, our model retains positional priors from the Grad-CAM heatmaps with almost identical structure to ground truth heatmaps. On the other hand, when the raw Grad-CAM heatmaps are inaccurate or very noisy (Row 4), our method can ignore the Grad-CAM heatmaps.

\setlength\extrarowheight{1pt}
\subsection{Ablation Studies}
In the ablation studies shown in \cref{tab:method_ablation}, we tested the contribution of the various components of our method. We also tested different parameters of the diffusion sampling process in \cref{tab:diffusion_ablation}. All ablation studies were performed using 10\% of the GazeFollow labels and the teacher models were trained only on supervised data (without MT) to simplify training. Additionally, in \cref{tab:suponly_results}, we analyze the effect of the VQA priors by training the baselines in fully-supervised settings with/without Grad-CAM heatmaps.

\begin{table}[t]
\parbox[t]{.48\linewidth}{
\caption{\textbf{Ablations of ways for generating the pseudo annotations.}}
\label{tab:method_ablation}
\centering
\begin{tabular}{l ccc}
\Xhline{2\arrayrulewidth}
\multirow{2}{*}{Method}       & \multicolumn{2}{c}{Dist. $\downarrow$}    & \multirow{2}{*}{AUC  $\uparrow$}                         \\
                                                               & Avg.                      & Min.   &                   \\ \Xhline{2\arrayrulewidth}
  No Grad-CAM              & 0.191                    & 0.125          & 0.867                                          \\
 No Refine                  & 0.237                     & 0.166        & 0.833                                         \\                          
 Argmax Refine             & 0.207                     & 0.139         & 0.869                                          \\ 
 Direct Mapping          & 0.190                     & 0.122            & 0.878                                          \\
   \Xhline{\arrayrulewidth}
    Proposed GCDR  & \textbf{0.179} & \textbf{0.115} & \textbf{0.886} \\ \Xhline{2\arrayrulewidth}
\end{tabular}
}
\hfill
\parbox[t]{.48\linewidth}{
\centering
\caption{\textbf{Ablations of parameters of the diffusion model refining process.}}
\label{tab:diffusion_ablation}
\begin{tabular}{c c | ccc}
\Xhline{2\arrayrulewidth}
\multirow{2}{*}{Noise.$t$} & \multirow{2}{*}{Inf. Steps}       & \multicolumn{2}{c}{Dist. $\downarrow$}    & \multirow{2}{*}{AUC  $\uparrow$}                         \\
                         &                                      & Avg.                      & Min.   &                   \\ \Xhline{2\arrayrulewidth}
  300    & 2        & 0.186                    & 0.120          & 0.877                                          \\
 200 &  2                 & 0.184                     & 0.117        & 0.883                                         \\                          
 250 & 5             & 0.180                     & \textbf{0.115}         & 0.883                                         \\ 
 250 & 2 & \textbf{0.179} & \textbf{0.115} & \textbf{0.886} \\ \Xhline{2\arrayrulewidth}
\end{tabular}
}
\end{table}

\subsubsection{Refinement methods}
In \textit{No Grad-CAM}, the diffusion model samples pseudo annotations from Gaussian noise without using Grad-CAM heatmaps. Without the VQA model priors, there is a significant performance decrease. \textit{No Refinement} directly uses the raw Grad-CAM heatmaps as pseudo annotations without refinement, which also leads to a significant performance decrease due to the noisy nature of the Grad-CAM heatmaps. \textit{Argmax Refinement} generates a Gaussian heatmap from the maximum point of the Grad-CAM heatmap as pseudo annotation. Despite the improvement over \textit{No refinement} due to the ``cleaner" annotation pattern, performance is still far behind our method, because some of the Grad-CAM heatmaps are low-quality  (\Fref{fig:qualitative_vis}), so Gaussian ends up on outliers or incorrect locations. On the contrary, our method can ignore the Grad-CAM heatmaps in these cases. In \textit{Direct Mapping}, we trained a U-Net to learn a ``direct mapping'' from Grad-CAM heatmaps to groundtruth heatmaps which we then applied to the unlabeled data. This shows a large performance decrease, because it cannot capture the full distribution that a diffusion model learns. 

All the above ablations validate the importance of both the Grad-CAM heatmaps priors and the priors that the diffusion model learns. 

\subsubsection{Added Noise Magnitude Effects}\label{sec:test_noise}

In this section, we analyze the effect of the added noise magnitude on the diffusion output. As shown in \cref{tab:diffusion_ablation}, adding noise at the 250th step achieved the best trade-off between the VQA prior and the pretrained annotation prior. As a reminder, the later the timestep we add the noise, the larger its magnitude (the more the model ignores the Grad-CAM).

\begin{figure}[t]
\setlength\tabcolsep{0.5pt}
\def\arraystretch{0.50}
\centering
\begin{tabular}{ccccc}
    Grad-CAM &
  $t = 100$ &
 $\bm{t = 250}$ &
 $t = 400$ &
 Pure noise\\
 \includegraphics[width=0.18\linewidth, height=0.6in]{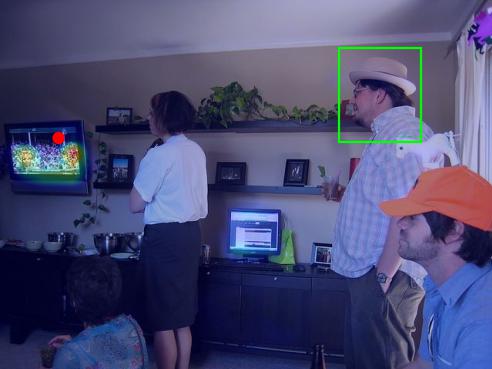}&
 \includegraphics[width=0.18\linewidth, height=0.6in]{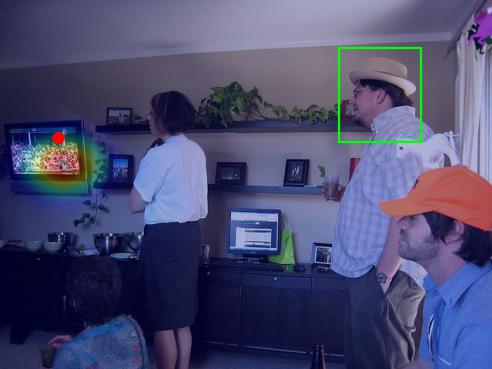} &
 \includegraphics[width=0.18\linewidth, height=0.6in]{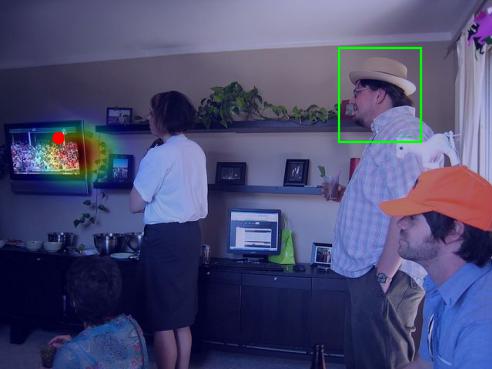} &
 \includegraphics[width=0.18\linewidth, height=0.6in]{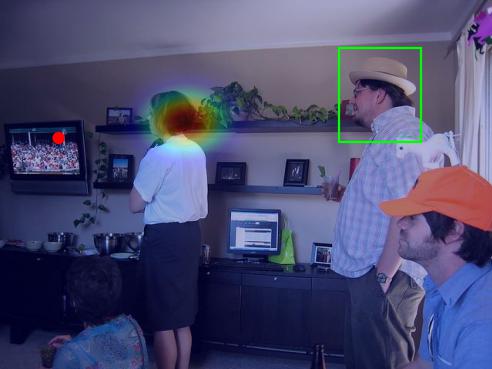} &
 \includegraphics[width=0.18\linewidth, height=0.6in]{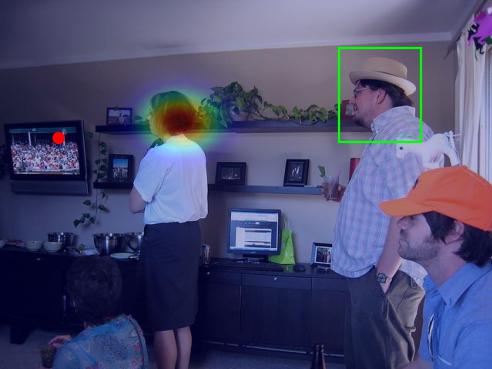} \\
 \includegraphics[width=0.18\linewidth, height=0.6in]{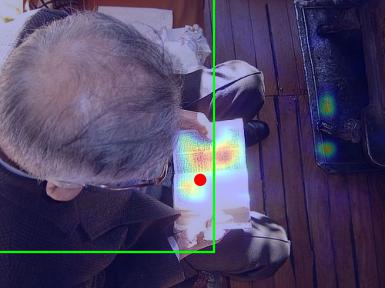} &
 \includegraphics[width=0.18\linewidth, height=0.6in]{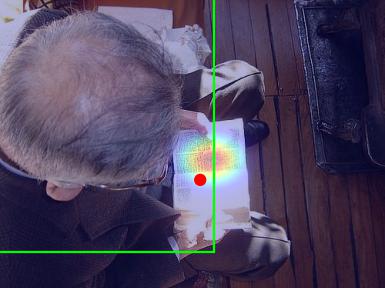} &
 \includegraphics[width=0.18\linewidth, height=0.6in]{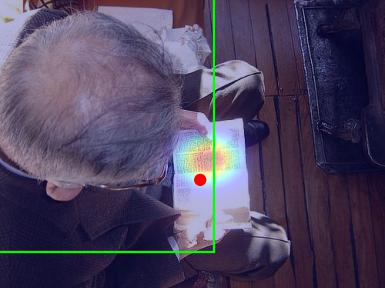} &
 \includegraphics[width=0.18\linewidth, height=0.6in]{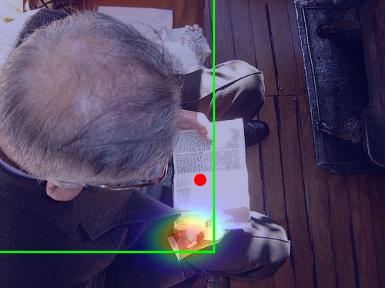} &
 \includegraphics[width=0.18\linewidth, height=0.6in]{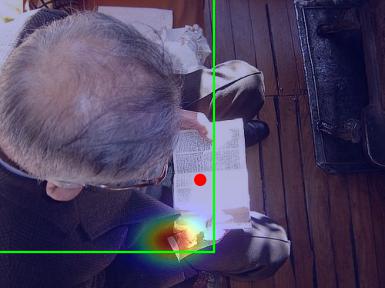}  \\
 \includegraphics[width=0.18\linewidth, height=0.6in]{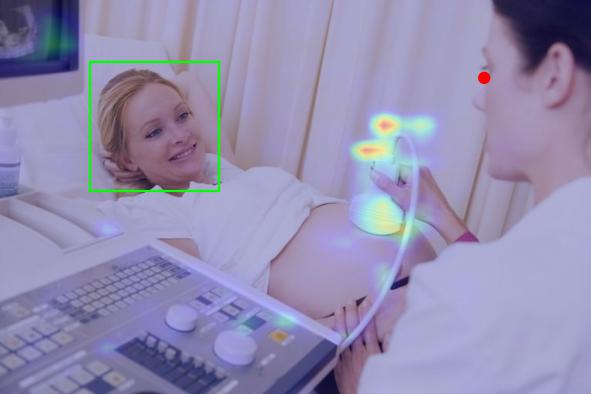} &
 \includegraphics[width=0.18\linewidth, height=0.6in]{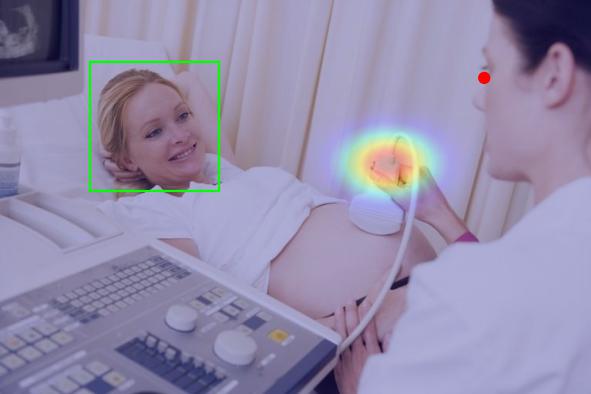} &
 \includegraphics[width=0.18\linewidth, height=0.6in]{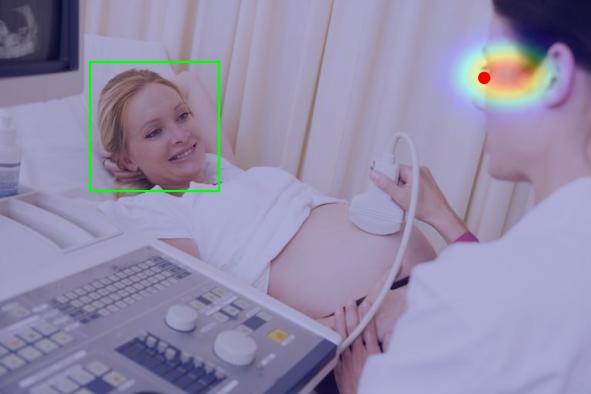} &
 \includegraphics[width=0.18\linewidth, height=0.6in]{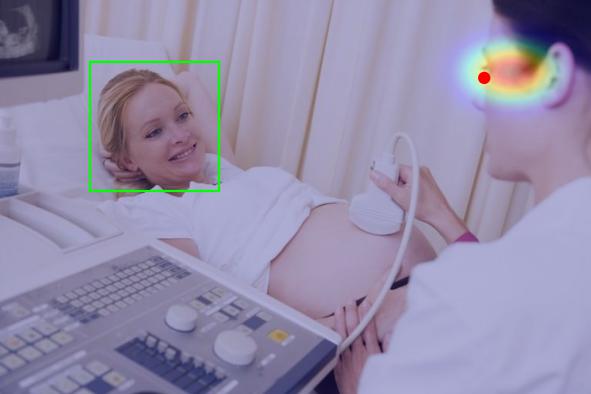} &
 \includegraphics[width=0.18\linewidth, height=0.6in]{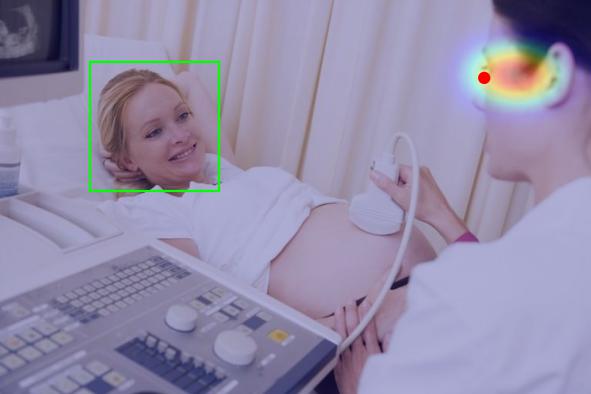} \\
  \includegraphics[width=0.18\linewidth, height=0.6in]{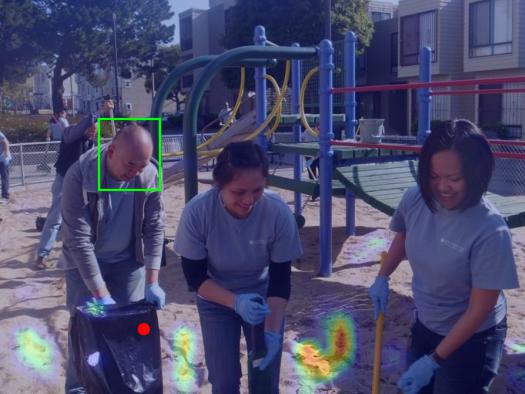} &
 \includegraphics[width=0.18\linewidth, height=0.6in]{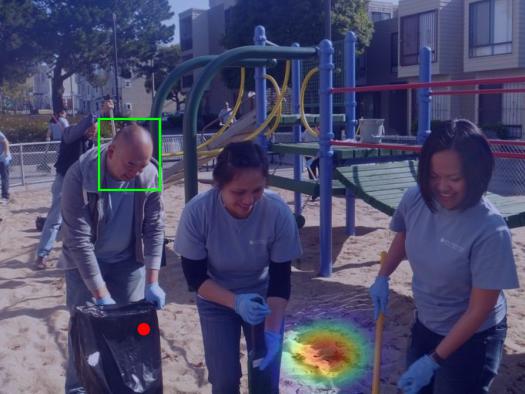} &
 \includegraphics[width=0.18\linewidth, height=0.6in]{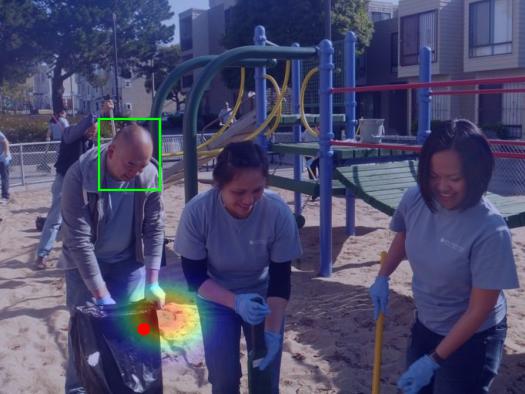} &
 \includegraphics[width=0.18\linewidth, height=0.6in]{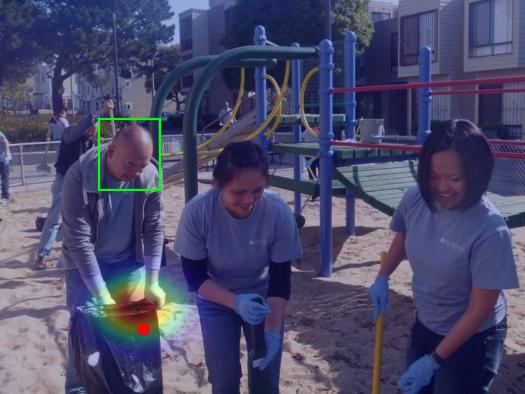} &
 \includegraphics[width=0.18\linewidth, height=0.6in]{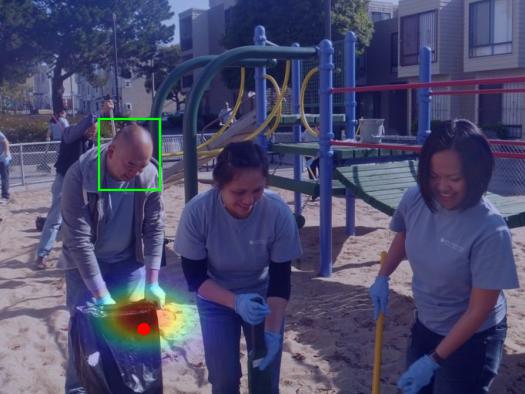} 
\end{tabular}
\caption{ \textbf{Diffusion model output for noise added at different timesteps}. Red dots represent the ground truth annotation. Adding noise at earlier steps generates outputs on high Grad-CAM response regions. Adding noise at later steps generates outputs similar to sampling from pure noise. Noise at step 250 is the best trade-off.}
\label{fig:compare_noise}
\end{figure}

We further visualized the outputs from the same diffusion model when adding noise at different time steps in \Fref{fig:compare_noise}. In the top 2 rows, the Grad-CAM heatmaps highlight correct locations, while in the bottom 2 rows, they are noisy or highlight incorrect locations. In all these cases, when adding noise at the 100th step, the outputs are located on the highlighted region of the Grad-CAM heatmap. When adding noise at the 400th step, the large magnitude of added noise leads to outputs similar to sampling directly from Gaussian noise without using the Grad-CAM heatmaps. Adding noise at step 250 appears to be the best trade-off. 

\begin{table}[h]
 \caption{\textbf{Results of using Grad-CAM heatmaps as a direct input}. When Grad-CAM heatmaps are directly input into the gaze inference process, both the VAT and the diffusion model show significant improvement in performance.}
\centering
\setlength\tabcolsep{0.7pt}
\begin{tabular}{l|l|ccc|ccc|ccc}
\Xhline{2\arrayrulewidth}
\multirow{3}{*}{Method} &\multirow{3}{*}{\makecell{Auxiliary\\Input}} & \multicolumn{3}{c|}{5\% labels}                                            & \multicolumn{3}{c|}{10\% labels}                                            & \multicolumn{3}{c}{20\% labels}                                          \\ \cline{3-11} 
  &  &  \multicolumn{2}{c}{Dist. $\downarrow$}  &  \multicolumn{1}{c|}{\multirow{2}{*}{AUC $\uparrow$}}           & \multicolumn{2}{c}{Dist $\downarrow$}                 & \multicolumn{1}{c|}{\multirow{2}{*}{AUC $\uparrow$}}          & \multicolumn{2}{c}{Dist. $\downarrow$}                  & \multirow{2}{*}{AUC $\uparrow$} \\
        &   &                           \multicolumn{1}{l}{Avg.} & \multicolumn{1}{l}{Min.} &    \multicolumn{1}{c|}{}       & \multicolumn{1}{l}{Avg.} & \multicolumn{1}{l}{Min.} & \multicolumn{1}{c|}{}  & \multicolumn{1}{l}{Avg.} & \multicolumn{1}{l}{Min.} & \\ \Xhline{2\arrayrulewidth}         
VAT & None   &   0.230 & 0.161 & 0.835                         & 0.202                    & 0.133   & 0.869                                   & 0.182                    & 0.116   & 0.892                                     \\
VAT-GC & Grad-CAM     &   0.210 &  0.144 & \textbf{0.848}                  & 0.188                    & 0.121      & \textbf{0.884}                               & 0.173                    & 0.107        & \textbf{0.898}                              \\
Diffusion & None     &      0.230 & 0.160 & 0.768           & 0.203                   &  0.135       &   0.847                          &    0.189                &  0.124      &    0.849                
 \\
Diffusion-GC & Grad-CAM         & \textbf{0.199}                    & \textbf{0.131}     & 0.803              & \textbf{0.177}                    & \textbf{0.112}     & 0.846                                & \textbf{0.167}                    & \textbf{0.103}     & 0.870         
 \\ \Xhline{2\arrayrulewidth}
\end{tabular}
 \label{tab:suponly_results}
\end{table}

\subsubsection{The Effect of the Grad-CAM Heatmaps} \label{sec:fullysup}
In this section, we test how much guidance signal is provided by the VQA priors. This cannot be directly tested in the semi-supervised scenario, so we tested by using the Grad-CAM heatmaps as a direct input to a fully supervised gaze following model. We trained the teacher models of Sec. \ref{sec:semi_results} using different amounts of labeled data only, and introduce a new baseline: {\it Diffusion-GC}, where the diffusion model trained on labeled images was used to refine the Grad-CAM heatmaps computed directly on the test set. Note that {\it VAT-GC} and {\it Diffusion-GC} require the Grad-CAM heatmaps of the test images during inference, which may not be feasible in an online setting as it necessitates access to a large vision-language model.

~\cref{tab:suponly_results} presents the results. {\it VAT-GC} outperforms VAT in all cases, and {\it Diffusion-GC} performs best in the distance metrics. Both the VAT and diffusion model show large performance boosts when Grad-CAM heatmaps are directly used as input, which offers evidence that the VQA Grad-CAM heatmaps bring strong prior knowledge. As illustrated in \cref{sec:semi_results}, the lower AUC for {\it Diffusion} and {\it Diffusion-GC} can be attributed to the AUC on GazeFollow being evaluated with group-level annotations and favoring heatmaps spanning larger areas. We analyzed this in more detail in the supplementary material.

\subsection{Semi-supervised Gaze Following for Video}\label{adaption}

In this experiment, we finetune a pretrained image gaze following model to specific video scenes in a semi-supervised manner. 
We experiment on 10 TV shows from the VideoAttentionTarget dataset \cite{chong2020detecting}, comprising a total of 31,978 gaze annotations. For each video, we randomly select a clip that contains about 100 annotated frames (2--23\% of the entire video, mostly below 10\%) as groundtruth data. The remainder of the videos are regarded as unlabeled. Following \cite{chong2020detecting}, we started with the image version of the teacher model pre-trained on the GazeFollow dataset and extended it temporally through an additional ConvLSTM network. The model was then fine-tuned on the given annotations and the generated pseudo-annotations for each video. The experiment simulates a real-world scenario where only a few frames of a specific scene are annotated. 

\begin{figure}[t]
    \centering
    \includegraphics[width=\linewidth]{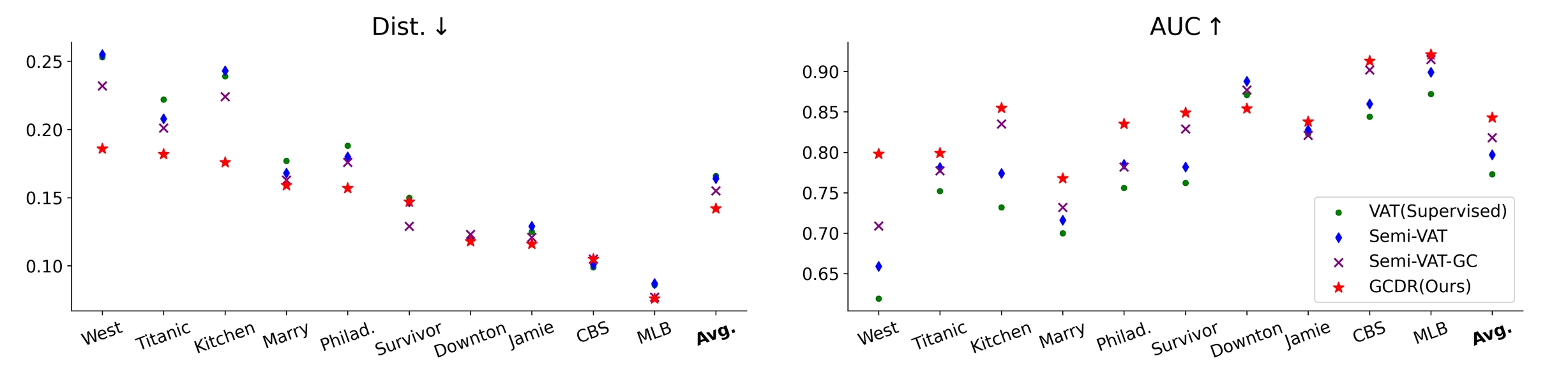}
  \caption{\textbf{Applying gaze following to videos in a semi-supervised manner}. GCDR achieves the overall best performance (Dist. and AUC averaged across videos: GCDR: 0.142, 0.843; Semi-VAT-GC: 0.155, 0.818; Semi-VAT: 0.164, 0.797; VAT: 0.166, 0.773). } 
  \label{fig:video_results_ten_updown}
\end{figure}

\Fref{fig:video_results_ten_updown} presents the video adaptation experiments results on each of the 10 videos. Our method outperforms other pseudo-annotation generation methods in both metrics on almost all videos.  When the baseline models do not perform well, the improvement is more pronounced (2\%--5\% in both metrics, more details in supplementary). These findings demonstrate that our method can be used effectively in video gaze following, needing to initially label only a few frames. 

\section{Conclusion}

We have proposed the first approach for generating high-quality pseudo-labels for the semi-supervised gaze following task. We leverage the priors from large VL models by computing Grad-CAM heatmaps from a pretrained VQA model that is prompted with a gaze following question. The Grad-CAM heatmaps offer strong guidance to the gaze target, but can be noisy. This led to a novel diffusion-based refinement method that refines these initial pseudo annotations with an annotation prior. Our approach works well on both image and video gaze following tasks with significant savings in the annotation effort. We hope our method will lead to the collection of larger gaze following datasets with annotation efforts similar to current datasets. We plan to apply diffusion-based refinement to ``noisy'' annotations generated by VL models, in new semi-supervised tasks.\\

%
%
\bibliographystyle{splncs04}
\bibliography{main}
\newpage

\renewcommand\thesection{S\arabic{section}}
\renewcommand\thefigure{S\arabic{figure}}
\renewcommand\thetable{S\arabic{table}}

\title{Diffusion-Refined VQA Annotations for Semi-Supervised Gaze Following: Supplementary Material} 

\titlerunning{GCDR-Gazefollowing}
\author{}
\authorrunning{Q. Miao et al.}
\institute{}
\maketitle

\begin{abstract}
In this supplementary material, we provide more details about the experiment settings of the main paper (\cref{train_detail}), we describe the baseline implementation details (\cref{baseline_detail}), we illustrate the details of the Grad-CAM heatmap generation procedure (\cref{gradcam_generation}), we present the results for each video in the video semi-supervised learning task (\cref{videoadapt_detail}), we discuss the reason for the lower AUC of the diffusion models in the fully supervised experiments (\cref{explain_auc}), we show the result of training with 50\% annotations on GazeFollow (\cref{50percent}) and training with 10\% labels on VideoAttentionTarget using the normal train-test split (\cref{videoatt_10percent}), we applied our methods to other gaze following models (\cref{semi_miao}), we analyzed the effect of the amount of unlabeled data by fixing the amount of labeled data (\cref{results_varyunlabel}), we visualize outputs from the diffusion model at intermediate inference steps (\cref{vis_step}), we provide detailed discussions on the available semi-supervised learning methods and their relations to gaze following (\cref{semisup_discussion}) and evaluate the raw Grad-CAM heatmaps when obtained with different extraction methods (\cref{ablation_gradcam}).
\end{abstract}

\section{Training Details} \label{train_detail}
Here we provide more details of the training parameters for the fully-supervised and semi-supervised settings on the GazeFollow dataset as well as semi-supervised training on the VideoAttentionTarget dataset. 

On the GazeFollow dataset, when training the teacher models with supervised data only, the batch size was set at 48 for all models (VAT, VAT-GC, and Diffusion model). The learning rate for training VAT and VAT-GC was $2.5\times 10^{-4}$. For the training of the diffusion model, we used a learning rate of $5\times 10^{-5}$. During the evaluation of the fully-supervised diffusion models in the ablation studies, we averaged 10 predictions from the diffusion model for each image to account for the variability introduced by the stochastic sampling process. In semi-supervised training, we used a learning rate of $2.5\times 10^{-4}$, with a decay factor of 0.2 at the 15th epoch when training with 5\%, 10\% and 20\% of annotations, and 25th epoch when training with 50\% of annotations. We randomly sampled 20 samples from the pool of labeled data in each batch when training with 5\%, 10\%, and 20\% of annotations to stabilize the training, following the training procedure of Mean Teacher \cite{meanteacher}. When refining the Grad-CAM heatmaps, we added noise at $t = 250$ when training with 10\%, 20\% and 50\% annotations, and $t = 200$ when training with 5\% annotations due to the weaker annotation prior when training the diffusion model with less labeled data.

In the task of semi-supervised training on videos, when training the teacher models, we tuned the learning rates for each video specifically based on the training loss, ranging from $5\times 10^{-6}$ to $5\times 10^{-5}$. When training the student model in semi-supervised training, for each video, we set the same learning rate for all teachers. The learning rates range from $1 \times 10^{-5}$ to $5 \times 10^{-5}$. 

\section{Baseline Implementation Details} \label{baseline_detail}

\begin{figure}[t]
\centering
\includegraphics[width=0.45\linewidth]{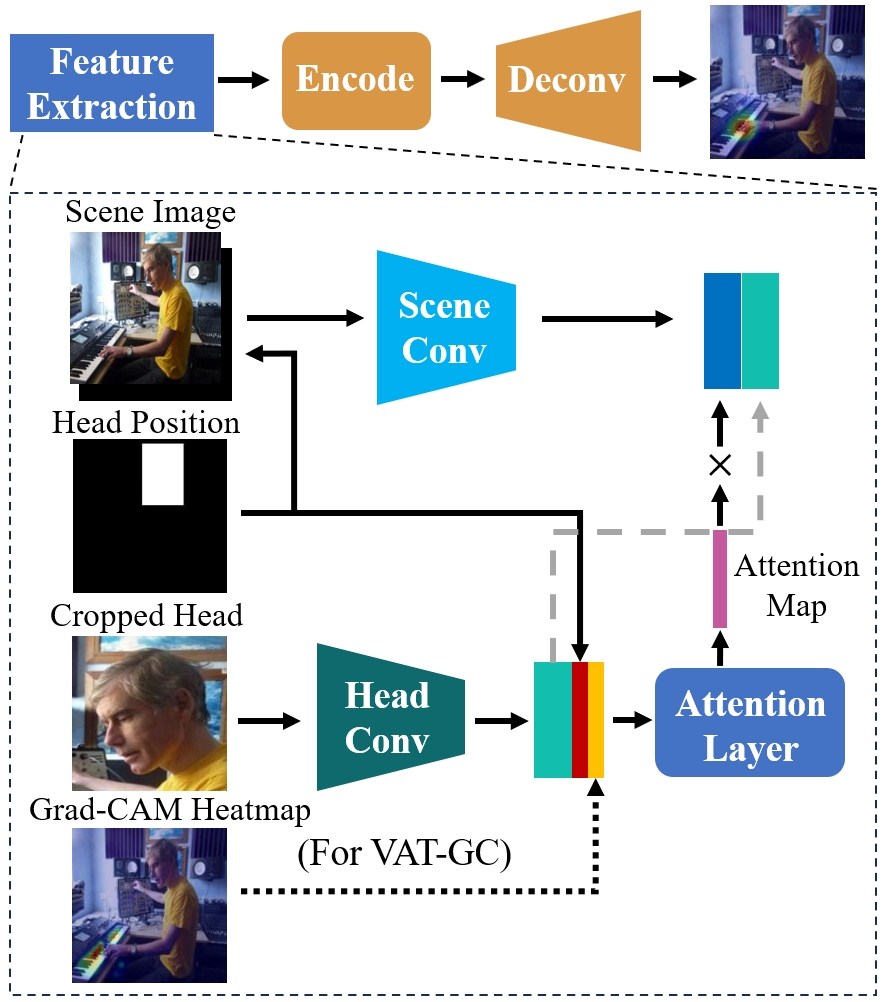}
\caption{\textbf{Structure of the VAT and VAT-GC baselines.} The model consists of a feature extraction module, an encoder, and a deconvolution module to output the gaze target heatmap. The figure shows the model's structure. For video gaze following, a Conv-LSTM layer is added between the encoder and deconvolution module. For VAT-GC, the Grad-CAM heatmap is concatenated with the extracted head feature to serve as conditional input.}
\label{fig:baseline}
\end{figure}

In this section, we provide the implementation details of the baseline teacher models.
VAT\cite{chong2020detecting} is a popular gaze following model that achieves close-to-human performance without using additional modalities. The structure of the model is shown in Fig.\ref{fig:baseline}. The model consists of a feature extraction module, an encoder, and a deconvolution module that outputs the heatmap. The feature extraction module takes a scene image, a binary head position mask, and the cropped head of the person as input, and outputs the extracted features from two pathways for encoding the scene and gaze features. The original model consists of two output branches for gaze target heatmap prediction and in/out prediction. In our experiments, we only used the heatmap prediction branch as we were focused only on the gaze target prediction task. When training the diffusion model, we used the same feature extraction module as the VAT model to extract conditional features that are input to the denoiser U-Net.

For VAT-GC, we modified the head pathway of the VAT model to take the raw Grad-CAM heatmap as additional input, as shown in the dashed lines of Fig.\ref{fig:baseline}. The attention layer was also modified according to the new input dimension.

For VAT-MT, we used the Mean Teacher method \cite{meanteacher} to train the VAT model on both labeled and unlabeled data. The student VAT model is trained with a combination of an L2 loss on labeled data, and a consistency loss between the outputs from the student and teacher model on all data. We used KL-Divergence as the consistency loss. We trained GCDR-MT in a similar manner by using the diffusion model as the student model.

\section{Details of Grad-CAM Heatmap Generation}\label{gradcam_generation}
In this section, we provide a detailed description of our Grad-CAM heatmap extraction procedure using the OFA model \cite{wang2022ofa}. The OFA model has a transformer-based encoder-decoder structure. Given an image with the target person overlaid by the detected bounding box and the gaze following question $Q$, it first uses a ResNet \cite{resnet} to extract visual features, which are then flattened into $N_v$ patch tokens: $\mathcal{V}=\{\bm{v}_{i} \in R^d\}_{i=1}^{N_v}$. Meanwhile, it uses byte-pair encoding (BPE) \cite{sennrich2015neural} to transform the text sequence of the question $Q$ into a subword sequence and then embed them into text tokens $\mathcal{Q}=\{\bm{q}_{j} \in R^d\}_{j=1}^{N_q}$. The visual and text tokens are then concatenated and fed into the transformer encoder. The transformer decoder takes in the output embeddings from the encoder, and generates output tokens auto-regressively, starting with a specific $\mathrm{[BOS]}$ token as the query.

After getting the output tokens from the VQA model, we compute the Grad-CAM heatmap $\bm{g}$ on the decoder cross-attention weights between the last input query token, and the image patch tokens when generating the word-of-interest during the auto-regressive generation process. In most cases, the answer is just a single noun (e.g., {\it``food''} or {\it ``plant''}). In such cases, we directly compute the Grad-CAM heatmap when the VQA model outputs this specific token. When the answer contains multiple words, such as {\it ``The girl on the left''}, we use parts-of-speech (POS) tagging using spaCy \cite{honnibal2020spacy} to find the word of interest. Specifically, each word is assigned a POS label, and only the words with a label of either noun or proper noun are selected \cite{shen2023text}. When the answer contains multiple nouns, such as {\it ``The boy wearing glasses''}, we select the first noun as it is the representative word in most cases. Finally, the Grad-CAM heatmaps are linearly interpolated to have the same size as the ground truth heatmaps in the training set. 

\section{Detailed Results of the Video Experiment} \label{videoadapt_detail}
In this section, we provide detailed results of the semi-supervised learning experiments on VideoAttentionTarget, including the exact performance of all methods, as well as the full video names and the number of annotations used in each video. The results are shown in Tab.\ref{tab:video_adapt_detail}. 

\begin{table}[t]
\centering
\caption{\textbf{Performance of different methods for each video in semi-supervised finetuning.} We also show the number of annotations used for fine-tuning and the ratio of the visible annotations with respect to all annotations in the entire video. GCDR performs best on almost all videos in both metrics.}
\resizebox{\textwidth}{!}{%
\begin{tabular}{l|c|c|cc|cc|cc|cc}
\Xhline{2\arrayrulewidth}
\multirow{2}{*}{Video}                                                          & \multirow{2}{*}{\begin{tabular}[c]{@{}c@{}}\# Annot. \\ Visible\end{tabular}} & \multirow{2}{*}{\begin{tabular}[c]{@{}c@{}}\ Annot. \\ Ratio \end{tabular}} & \multicolumn{2}{c|}{VAT}      & \multicolumn{2}{c|}{Semi-VAT}       & \multicolumn{2}{c|}{Semi-VAT-GC}    & \multicolumn{2}{c}{GCDR}                 \\ \cline{4-11} 
                                                                                &                                                                               &                                                                             & \multicolumn{1}{c|}{Dist. $\downarrow$}   & AUC $\uparrow$ & \multicolumn{1}{c|}{Dist.$\downarrow$}   & AUC $\uparrow$ & \multicolumn{1}{c|}{Dist. $\downarrow$}   &  AUC $\uparrow$ & \multicolumn{1}{c|}{Dist. $\downarrow$}            &  AUC $\uparrow$        \\ \Xhline{2\arrayrulewidth}
\begin{tabular}[c]{@{}l@{}} \it{West World} \\ {}\end{tabular}                                                                      & 109                                                                           & 11.61\%                                                                        & \multicolumn{1}{c|}{0.253} & 0.619 & \multicolumn{1}{c|}{0.255} & 0.659         & \multicolumn{1}{c|}{0.232} & 0.709         & \multicolumn{1}{c|}{\textbf{0.186}} & \textbf{0.798} \\ \hline 
\begin{tabular}[c]{@{}l@{}} \it{Titanic} \\ {}\end{tabular}                                                                        & 114                                                                           & 3.36\%                                                                        & \multicolumn{1}{c|}{0.222} & 0.752 & \multicolumn{1}{c|}{0.208} &  0.781         & \multicolumn{1}{c|}{0.201} & 0.777          & \multicolumn{1}{c|}{\textbf{0.182}} & \textbf{0.799} \\ \hline
\begin{tabular}[c]{@{}l@{}}{\it Hell's} \\ {\it Kitchen}\end{tabular}                       & 101                                                                           & 23.38\%                                                                         & \multicolumn{1}{c|}{0.239} & 0.732 & \multicolumn{1}{c|}{0.243} &  0.774         & \multicolumn{1}{c|}{0.224} &  0.835        & \multicolumn{1}{c|}{\textbf{0.176}} & \textbf{0.855} \\ \hline
\begin{tabular}[c]{@{}l@{}}{\it I Wanna} \\ {\it Marry Harry}\end{tabular}                  & 74                                                                           & 9.18\%                                                                        & \multicolumn{1}{c|}{0.177} & 0.700 & \multicolumn{1}{c|}{0.168} & 0.716 & \multicolumn{1}{c|}{0.163} & 0.732          & \multicolumn{1}{c|}{\textbf{0.159}} & \textbf{0.768}          \\ \hline
\begin{tabular}[c]{@{}l@{}}{\it It's Always Sunny} \\  {\it in Philadelphia}\end{tabular} & 116                                                                           & 5.02\%                                                                        & \multicolumn{1}{c|}{0.188} & 0.756 & \multicolumn{1}{c|}{0.180} &   0.785        & \multicolumn{1}{c|}{0.176} &  0.782         & \multicolumn{1}{c|}{\textbf{0.157}} & \textbf{0.835} \\ \hline
\begin{tabular}[c]{@{}l@{}} \it{Survivor} \\ {}\end{tabular}                                                                         & 73                                                                            & 3.99\%                                                                        & \multicolumn{1}{c|}{0.150} & 0.762  & \multicolumn{1}{c|}{0.147} & 0.782          & \multicolumn{1}{c|}{\textbf{0.129}} & 0.829 & \multicolumn{1}{c|}{0.147} & \textbf{0.849}          \\ \hline
\begin{tabular}[c]{@{}l@{}} {\it Downton} \\ {\it Abbey}\end{tabular}                         & 108                                                                           & 3.54\%                                                                      & \multicolumn{1}{c|}{0.121} & 0.871 & \multicolumn{1}{c|}{0.120} & \textbf{0.888}  & \multicolumn{1}{c|}{0.123} & 0.877 & \multicolumn{1}{c|}{\textbf{0.118}} & 0.854          \\ \hline
\begin{tabular}[c]{@{}l@{}}{\it Jamie} \\ {\it Oliver}\end{tabular}                                                                  & 70                                                                           & 6.30\%                                                                        & \multicolumn{1}{c|}{0.125} & 0.823 & \multicolumn{1}{c|}{0.129} &  0.828         & \multicolumn{1}{c|}{0.121} & 0.821 & \multicolumn{1}{c|}{\textbf{0.116}} & \textbf{0.838}          \\ \hline
\begin{tabular}[c]{@{}l@{}} {\it CBS This} \\ {\it Morning}\end{tabular}                     & 96                                                                            & 2.21\%                                                                     & \multicolumn{1}{c|}{0.099} & 0.844 & \multicolumn{1}{c|}{\textbf{0.101}} & 0.860 & \multicolumn{1}{c|}{0.105} & 0.902          & \multicolumn{1}{c|}{0.105} & \textbf{0.913}          \\ \hline

\begin{tabular}[c]{@{}l@{}} \it{MLB} \\ {\it Interview}\end{tabular}                        & 95                                                                            & 3.86\%                                                                       & \multicolumn{1}{c|}{0.086} & 0.872 & \multicolumn{1}{c|}{0.087} & 0.899          & \multicolumn{1}{c|}{0.077} & 0.915 & \multicolumn{1}{c|}{\textbf{0.076}} & \textbf{0.921}       
\\ \Xhline{2\arrayrulewidth}
\end{tabular}
}
\label{tab:video_adapt_detail}
\end{table}
As discussed in the main paper, our method performs the best on almost all videos. We would like to note that, in real-world applications when videos are collected from a new scene, users can compare the raw Grad-CAM heatmaps with the annotations from a few annotated frames. Based on the Grad-CAM heatmap quality of that specific scene, users can adjust the magnitude/timestep of the noise added to the Grad-CAM heatmap to decide the extent of information to be kept. In our method, we added noise at the 300th step instead of 250 for \textit{I Wanna Marry Harry} and \textit{Survivor} due to the lower quality of the Grad-CAM heatmaps for those videos. Overall, the results show that our method can also be applied to new videos, of which only a small number of frames are annotated.

\section{Diffusion Performance regarding AUC} \label{explain_auc}

\begin{figure}[t]
\centering
\begin{tabular}{cc}
 VAT-GC &
 Diffusion-GC\\
\includegraphics[width=0.26\linewidth]{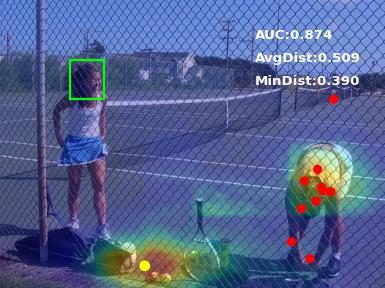}&
 \includegraphics[width=0.26\linewidth]{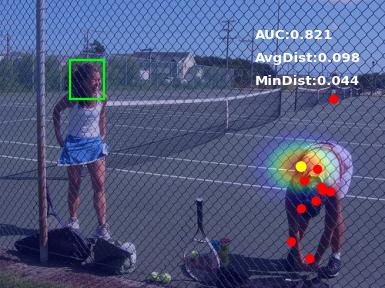} \\
\includegraphics[width=0.26\linewidth]{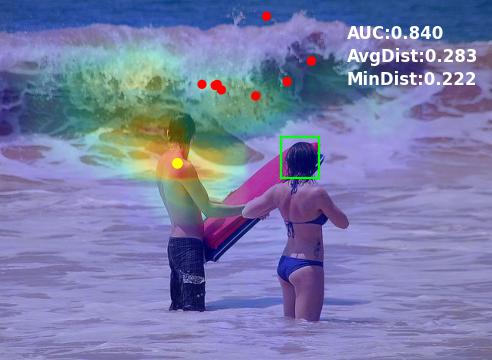}&
 \includegraphics[width=0.26\linewidth]{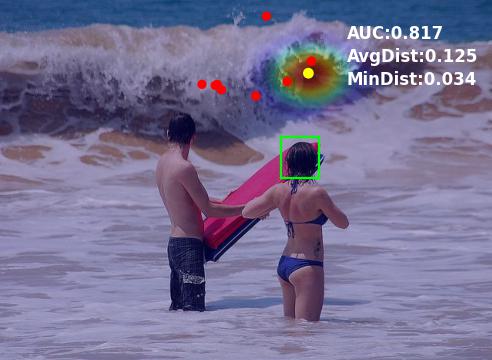} \\
\includegraphics[width=0.26\linewidth]{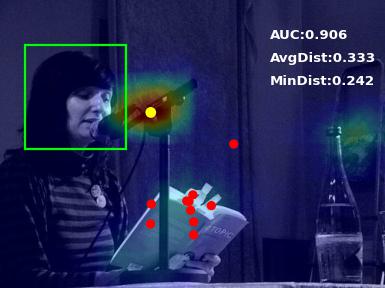}&
 \includegraphics[width=0.26\linewidth]{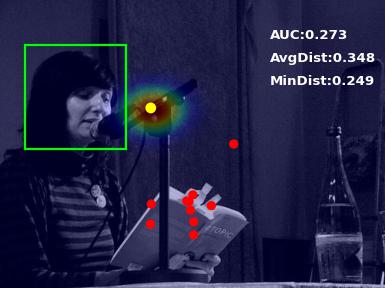} \\
\end{tabular}
\caption{\textbf{Example demonstrations of the lower AUC for diffusion model outputs on GazeFollow test images}. For each image, the predicted gaze target is shown as a yellow dot and the ground truth annotations are shown as red dots. The lower AUC for the diffusion model is mainly due to the smaller overlap between the output heatmaps and the annotations. The AUC is negatively affected when the annotations span a larger area, despite the predicted targets being closer to the actual gaze targets (Rows 1-2). In Row 3, although both models predict the same incorrect target, VAT-GC attains a much higher AUC due to the overlap of the group-level annotations with a weak response from the model.}
\label{fig:explain_auc}
\end{figure}

In this section, we visually illustrate the cause of the lower performance in the AUC metrics for the diffusion baselines in Tab. 4 in the main paper. In the evaluation of the AUC metric on the GazeFollow test set, for each image, the heatmap is evaluated with a binary map where each annotation from the 10 annotators is assigned as one. Therefore, the AUC favors heatmap predictions covering a larger area so that they overlap with more annotations, even if the heatmap predictions have high responses on incorrect locations, as shown in Fig.\ref{fig:explain_auc}. As the diffusion model learns to model the distribution of human-labeled annotation of the training data, which is a single Gaussian, the outputs of Diffusion-GC tend to overlap with fewer human annotations, even when the predicted target locations are more accurate (Rows 1-2). In Row 3, VAT-GC and Diffusion-GC show very similar predictions with the predicted targets on incorrect locations, while VAT-GC shows much higher AUC due to the overlap of a weak response with a cluster of annotations. 

These examples also serve as a clue for the smaller performance advantage in AUC compared to the distance metrics in the semi-supervised results in the main paper. As mentioned in the paper, our methods perform better in predicting the probable location of the target than in predicting the exact shape of the heatmap.

\section{Training with 50\% of Annotations} \label{50percent}

\begin{table}[h]
\caption{\textbf{Semi-supervised experiments when training with 50\% labels}} 
\centering
\begin{tabular}{l|ccc}
\Xhline{2\arrayrulewidth}
Method & \multicolumn{2}{c}{Dist. $\downarrow$}  &  \multirow{2}{*}{AUC $\uparrow$} \\
& \multicolumn{1}{l}{Avg.} & \multicolumn{1}{l}{Min.} & \\
\Xhline{2\arrayrulewidth}
VAT (Supervised) & 0.160 & 0.096   & 0.912 \\ \Xhline{\arrayrulewidth}   
Semi-VAT & 0.153 & 0.090  & 0.916 \\
Semi-VAT-GC & 0.149 & 0.086  & 0.917 \\
VAT-MT & 0.150 & 0.086 & 0.918 \\
GCDR (Ours) & 0.143 & 0.082  & 0.919 \\
GCDR-MT (Ours) & \textbf{0.141}  & \textbf{0.080}  & \textbf{0.920} \\ \Xhline{\arrayrulewidth}
VAT (100\% labels) & 0.137 & 0.077  & 0.921 \\ \Xhline{2\arrayrulewidth}
\end{tabular}
\label{tab:semi_50percent}
\end{table}

In \cref{tab:semi_50percent}, we provide the semi-supervised experiment results when training with 50\% annotations. Our method still outperforms the baselines. We achieved close performance to the VAT model trained with full annotations. The performance did not outperform the VAT trained with 100\% annotations because only 50\% unlabeled data can be used in this scenario, which is the same amount as labeled data, while semi-supervised learning usually assumes that the unsupervised data has a much larger scale than labeled data (As shown by results of training with 5\%, 10\% and 20\% in the paper). Based on the analysis results of the amount of unlabeled data in \cref{tab:semi_fix0.1}, we believe the performance of our method can outperform the model trained with 100\% annotations when additional large unsupervised data is available.

\section{Training with 10\% labels on VideoAttentionTarget} \label{videoatt_10percent}

In the main paper, we showed the results of fine-tuning the VAT model on each of the 10 videos in VideoAttentionTarget in a semi-supervised manner. Here we adopt an ordinary train-test evaluation scheme and evaluate the performance of the VAT model trained with 10\% of annotations in the training set and evaluated on the test set of VideoAttentionTarget, with the pseudo annotations of the unlabeled data generated by different methods. \cref{tab:video_semi} shows that our method (GCDR) outperforms the pseudo annotation generation baselines in both metrics. Our method even slightly outperforms the VAT model trained with full labels on VideoAttentionTarget, possibly due to the stochastic nature of the diffusion model.

\begin{table}[h]
\centering
\setlength\tabcolsep{1.0pt}
\caption{\textbf{Results of Training with 10\% labels on VideoAttentionTarget}}
\begin{tabular}{l c c}
    \toprule
    Method & Dist. $\downarrow$  & AUC $\uparrow$ \\
    \toprule
    VAT (Supervised) & 0.144 & 0.825\\
    Semi-VAT & 0.136 & 0.844\\
    Semi-VAT-GC & 0.141 & 0.856\\
    GCDR & \textbf{0.127} & \textbf{0.865}\\
    \Xhline{\arrayrulewidth} 
    VAT (100\% labels) & 0.134 & 0.860 \\
    \toprule
    \end{tabular}
    \label{tab:video_semi}
\end{table}

\section{Experiments with Other Gaze Following Models} \label{semi_miao}
In the main paper, we based our experiments on VAT. Here we show that our method is also applicable to other gaze following models. In this section, we applied our semi-supervised learning method to Miao \etal \cite{miao2023patch} and Lian \etal \cite{lian2018believe}. Note that \cite{miao2023patch} uses depth as additional input and has another output branch for patch distribution prediction (PDP) of the gaze target. When applying our method to \cite{miao2023patch}, we modified the feature extraction module of the diffusion model to be the same as \cite{miao2023patch} to use depth as input. In semi-supervised training, we computed the patch-level gaze distribution following \cite{miao2023patch} from the generated pseudo heatmaps as the pseudo label for the PDP branch.

We tested the pseudo annotation generation methods with 10\% of annotations. Similar to Tab.1 in the paper, we built {\it Semi-Miao} and {\it Semi-Lian} by training \cite{miao2023patch} and \cite{lian2018believe} on the labeled data to generate pseudo labels. We also built {\it Semi-Miao-GC} and {\it Semi-Lian-GC} by modifying the models to use the Grad-CAM heatmap as conditional input. {\it Miao-MT} was implemented by applying the consistency loss on the predicted heatmaps and patch-level distributions in \cite{miao2023patch} from the teacher and student models, while {\it Lian-MT} applies consistency loss between the predicted gaze directions and heatmaps from the teacher and student models. We still have two versions of our method: {\it GCDR} and {\it GCDR-MT}, except that when applied to \cite{miao2023patch}, the feature extraction module of the diffusion model was the same as \cite{miao2023patch}.

\setlength\tabcolsep{1pt}
\begin{table}[h]
\vspace{-0.5cm}
\begin{minipage}[t]{.46\linewidth}
\centering
\caption{\footnotesize \textbf{Using \cite{miao2023patch} as base model}} 
\footnotesize
\centering
\begin{tabular}{l|ccc}
\Xhline{2\arrayrulewidth}
Method & \multicolumn{2}{c}{Dist. $\downarrow$}  &  \multirow{2}{*}{AUC $\uparrow$} \\
& \multicolumn{1}{l}{Avg.} & \multicolumn{1}{l}{Min.} & \\
\Xhline{2\arrayrulewidth}
Miao (10\% labels) & 0.190 & 0.123   & 0.877 \\ \Xhline{\arrayrulewidth}   
Semi-Miao & 0.180 & 0.115  & 0.892 \\
Semi-Miao-GC & 0.185 & 0.117  & 0.893 \\
Miao-MT & 0.217 & 0.146 & 0.854 \\
GCDR (Ours) & 0.167 & 0.104  & 0.897 \\
GCDR-MT (Ours) & \textbf{0.163}  & \textbf{0.100}  & \textbf{0.900} \\ \Xhline{\arrayrulewidth}
Miao (20\% labels) & 0.166 & 0.103  & 0.900 \\ \Xhline{2\arrayrulewidth}
\end{tabular}
\label{tab:semi_pdp}
\end{minipage}
\hfill
\begin{minipage}[t]{.46\linewidth}
\caption{\footnotesize \textbf{Using \cite{lian2018believe} as base model}} 
    \footnotesize
    \centering
    \begin{tabular}{l|ccc}
    \Xhline{2\arrayrulewidth}
    Method & \multicolumn{2}{c}{Dist. $\downarrow$}  &  \multirow{2}{*}{AUC $\uparrow$} \\
    & \multicolumn{1}{l}{Avg.} & \multicolumn{1}{l}{Min.} & \\
    \Xhline{2\arrayrulewidth}
    Lian (10\% labels) & 0.216 & 0.145   & 0.875 \\ \Xhline{\arrayrulewidth}   
    Semi-Lian & 0.203 & 0.133  & 0.886 \\
    Semi-Lian-GC & 0.198 & 0.129  & 0.889 \\
    Lian-MT & 0.196 & 0.128 & 0.882 \\
    GCDR (Ours) & 0.178 & 0.112  & 0.895 \\
    GCDR-MT (Ours) & \textbf{0.171}  & \textbf{0.106}  & \textbf{0.898} \\ \Xhline{\arrayrulewidth}
    Lian (20\% labels) & 0.186 & 0.120  & 0.892 \\ \Xhline{2\arrayrulewidth}
    \end{tabular}
\label{tab:semi_lian}
    
    \end{minipage}
    \vspace{-0.3cm}
\end{table}

\cref{tab:semi_pdp} and \cref{tab:semi_lian} present the results. Our method shows consistent improvements compared to the baseline methods, especially in the distance metrics. Our semi-supervised methods show comparable or better performance to the supervised model trained with 20\% annotations, the same as the results in the main paper. The Mean Teacher method failed when directly applied to \cite{miao2023patch}, possibly due to the instability when applying the consistency loss to outputs from two branches in \cite{miao2023patch}. This substantiates the effectiveness of our method when applied to other gaze following models, including models using additional modalities, by modifying the feature extraction module of the diffusion model to be the same as the base model for using the additional modalities.

\section{Effect of the Amount of Unlabeled data} \label{results_varyunlabel}

In the main paper, we ran experiments with different amounts of annotations in GazeFollow and treated the rest of the data as unlabeled. Though this experiment setting was designed following previous semi-supervised learning works \cite{NIPS2015_ec895663, chong2020detecting, lian2018believe}, the amount of both labeled and unlabeled data was different across settings. In this section, we specifically tested the effect of the amount of unlabeled data by fixing the number of labeled annotations as 10\%, and varying the amount of unlabeled data.

Results are shown in \cref{tab:semi_fix0.1}. Along with the results of training with 10\% labeled and 90\% unlabeled data copied from Tab.1,  we experimented using 10\% and 50\% unlabeled data in training. Our method shows consistent improvement over the baselines, and more unlabeled data leads to better performance.

\begin{table}[h]
\caption{\textbf{Training on 10\% labeled data and varying amounts of unlabeled data.}} 
\centering
\begin{tabular}{l|ccccccccc}
\Xhline{2\arrayrulewidth}
 Method & \multicolumn{3}{c|}{10\% unsupervised}                                            & \multicolumn{3}{c|}{50\% unsupervised}            & \multicolumn{3}{c}{90\% unsupervised}                                        \\ \cline{2-10} 
                & \multicolumn{2}{c}{Dist. $\downarrow$}  &  \multicolumn{1}{c|}{\multirow{2}{*}{AUC $\uparrow$}}           & \multicolumn{2}{c}{Dist $\downarrow$}                 & \multicolumn{1}{c|}{\multirow{2}{*}{AUC $\uparrow$}}          & \multicolumn{2}{c}{Dist. $\downarrow$}                  & \multirow{2}{*}{AUC $\uparrow$} \\
                                           & \multicolumn{1}{l}{Avg.} & \multicolumn{1}{l}{Min.} &    \multicolumn{1}{c|}{}       & \multicolumn{1}{l}{Avg.} & \multicolumn{1}{l}{Min.} & \multicolumn{1}{c|}{}  & \multicolumn{1}{l}{Avg.} & \multicolumn{1}{l}{Min.} & \\ \Xhline{2\arrayrulewidth} 
VAT (Supervised)                               & 0.202                    & 0.133  & \multicolumn{1}{c|}{0.869}              & 0.202                & 0.133        & \multicolumn{1}{c|}{0.869}      & 0.202                & 0.133        & 0.869                    \\
\Xhline{\arrayrulewidth}                

Semi-VAT            & 0.199  &   0.130                & \multicolumn{1}{c|}{0.872}          &            0.196    &  0.129                   &  \multicolumn{1}{c|}{0.874}      & 0.195                    & 0.128           & 0.875                \\
Semi-VAT-GC                     & 0.200                    & 0.132   & \multicolumn{1}{c|}{0.872}                        & 0.197                    & 0.128    & \multicolumn{1}{c|}{0.877}   & 0.195                    & 0.127       & 0.879             \\
VAT-MT                   & 0.202                    & 0.133  & \multicolumn{1}{c|}{0.875}                        & 0.194                    & 0.125   & \multicolumn{1}{c|}{0.878}  & 0.189                    & 0.122           & 0.882    \\  
GCDR (Ours)   &    0.185    &    0.119 &  \multicolumn{1}{c|}{0.880}     &   0.181    &  0.117  & \multicolumn{1}{c|}{0.883} &  0.179    &    0.115   &  0.886      \\ 
GCDR-MT (Ours)     &    \textbf{0.183}    &    \textbf{0.117} &  \multicolumn{1}{c|}{\textbf{0.881}}   &   \textbf{0.176}     &  \textbf{0.112}  & \multicolumn{1}{c|}{\textbf{0.886}}    & \textbf{0.172}    &   \textbf{0.108} &   \textbf{0.892}         \\ 
\Xhline{2\arrayrulewidth}  
\end{tabular}
\label{tab:semi_fix0.1}
\end{table}

\section{Visualizations of Diffusion Model Output in each Inference Step} \label{vis_step}

\begin{figure*}[t]
\centering
\includegraphics[width=\linewidth]{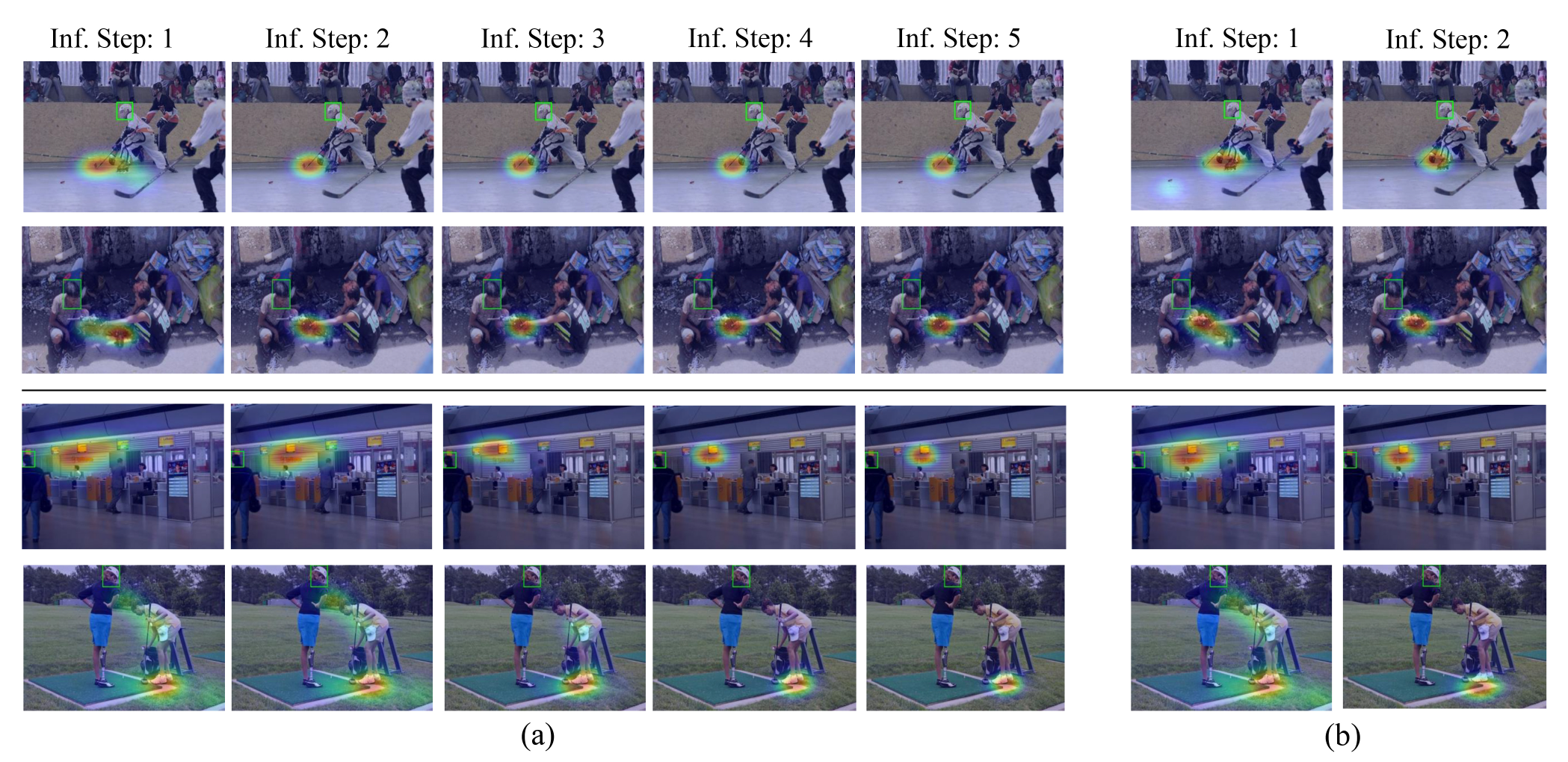}
\caption{\textbf{Visualizations of the diffusion model output in each step.} (a): 5 inference steps in total (b): 2 inference steps in total. The top 2 rows show the output during refinement of the Grad-CAM heatmaps, and the bottom 2 rows show the output when sampling from pure Gaussian noise. The diffusion model tends to generate more concentrated heatmaps during the inference process.}
\label{fig:vis_steps}
\end{figure*}

In our ablation study in the main paper, we presented the performance for various initial refinement timesteps, while the outputs are the outputs from the last inference step. In this section, we investigate the model's behavior by visualizing the output heatmaps at each inference step. Recall that we predict $\vect{h}_0$ directly at each inference step.

Fig.\ref{fig:vis_steps} shows the heatmap visualizations during the Grad-CAM heatmap refinement (Rows 1-2) and sampling from pure Gaussian noise (Rows 3-4). In the first step, the diffusion model usually outputs uncertain heatmaps that span a large area. During the iterative inference process, the output heatmaps become more concentrated, similar to the single Gaussian shape of human-labeled annotations. During refinement, the diffusion model outputs more concentrated heatmaps at earlier steps, due to the smaller magnitude of noise added compared to sampling from pure Gaussian noise. From (a) and (b) we observe inference with 2 steps shows a similar final output with 5 steps, therefore we choose 2 inference steps to reduce inference time.

\section{Discussion on Existing Semi-Supervised Learning Works} \label{semisup_discussion}
 In this section, we provide a more detailed discussion of the existing semi-supervised learning methods and their potential applicability in gaze following. As mentioned in the main paper, most semi-supervised learning methods target recognition and segmentation tasks and usually involve task-specific operations \cite{semisup_survey, kalluri2019universal, berthelot2019mixmatch}. Some semi-supervised recognition methods convert the predicted multi-class distribution from the teacher models to a one-hot vector in recognition \cite{ lee2013pseudo,sohn2020fixmatch} by selecting the class with the highest predicted confidence, while some work uses a 'sharpening' operation to obtain a softer vector similar to one-hot \cite{berthelot2019mixmatch, hu2021simple}, or requires the predicted class label from the student model to find a stable sample \cite{ke2019dual}. Meanwhile, some segmentation methods also compute pixel-wise one-hot pseudo-labels in segmentation by selecting the most confident class \cite{wang2022semi}, or compare the similarities between the predicted classification distributions from the teacher and student models in a pixel-wise manner \cite{kalluri2019universal}. In addition, some earlier semi-supervised methods use Generative Adversarial Networks (GANs) or Variational AutoEncoders (VAEs) to learn a latent space in which features from different classes are better separated \cite{ springenberg2015unsupervised, salimans2016improved, kingma2014semi,paige2017learning}. These classification-specific operations cannot be easily adapted to gaze following, as gaze following predicts a dense spatial heatmap for the target instead of predicting the class of the input. 
 Even if the target heatmap is flattened to a vector by treating each pixel as one category, it is incorrect to convert the prediction to a one-hot vector as it ignores the spatial correlations between neighboring pixels compared to using the spatial heatmap as pseudo annotations. 

\begin{table}[t]
\caption{\textbf{Results of applying "one-hot" transformation on the baseline predictions.} We generated a Gaussian on the maximum response pixel on the predicted heatmaps as pseudo-annotations for the baseline methods (Semi-VAT-Gaussian and Semi-VAT-GC-Gaussian). Other results are copied from Tab.1 for reference. Experiments are performed by training with 10\% annotations.} 
\centering
\begin{tabular}{l|ccc}
\Xhline{2\arrayrulewidth}
Method & \multicolumn{2}{c}{Dist. $\downarrow$}  &  \multirow{2}{*}{AUC $\uparrow$} \\
& \multicolumn{1}{l}{Avg.} & \multicolumn{1}{l}{Min.} & \\
\Xhline{2\arrayrulewidth}
VAT (Supervised) & 0.202 & 0.133   & 0.869 \\ \Xhline{\arrayrulewidth}   
Semi-VAT & 0.195 & 0.128  & 0.875 \\
Semi-VAT-Gaussian & 0.576 & 0.489 & 0.777 \\
Semi-VAT-GC & 0.195 & 0.127 & 0.879 \\
Semi-VAT-GC-Gaussian & 0.195 & 0.127  & 0.863 \\
GCDR (Ours) & 0.179  & 0.115  & 0.886 \\
GCDR-MT (Ours) & \textbf{0.172}  & \textbf{0.108}  & \textbf{0.892} \\ \Xhline{2\arrayrulewidth}
\end{tabular}
\label{tab:baseline_gaussian}
\end{table}

 In \cref{tab:baseline_gaussian}, we also tried generating a Gaussian on the maximum response pixel from the predicted heatmaps of the baseline models as pseudo-annotation, which mimics the procedures of generating softer "one-hot" labels. We didn't perform this on VAT-MT, because the Mean Teacher method enforces consistencies between the original output from the teacher and student models, while the operation of generating a Gaussian will change the output. After generating a Gaussian as the pseudo-annotations, Semi-VAT-GC shows the same performance in Dist., but with a lower AUC as the Gaussian heatmap support now spans a smaller area than the original prediction. The Semi-VAT totally fails after this operation, possibly because without the Grad-CAM heatmap prior, the maximum responses usually locate in incorrect locations, which makes the training very unstable. In contrast, the original prediction may possibly overlap with the correct target due to the larger and uncertain heatmap response, even when the predicted target is incorrect.
 
 In addition, semi-supervised learning methods usually involve strong augmentations and enforce consistencies between the student and teacher model outputs from the strongly and weakly augmented views of the input. Some methods used CutOut \cite{sohn2020fixmatch, french2020semi}, CutMix \cite{kim2020structured}, shear or translation of images \cite{zhang2021flexmatch, cubuk2020randaugment},  masking out image patches \cite{assran2022masked}, Mixup of images and labels \cite{berthelot2019mixmatch, semi-vit} as the strong augmentations for training the semi-supervised model. Unfortunately, none of these augmentations can be applied to gaze following which requires an intact scene image for inferring the target, where the person and target must both be in the image. 

 On the other hand, semi-supervised learning was also applied in certain regression tasks, such as crowd-counting. However, the related methods still mostly involve task-specific operations, including auxiliary tasks such as segmentation of crowd on the images \cite{liu2020semi, meng2021spatial, zhu2023multi}, crowd density ranking on image patches \cite{crowdcounting_density}, training uncertainty prediction module on cropped patches \cite{li2023calibrating}, which makes these methods not directly applicable to gaze following.  

Despite the plenty of non-applicable methods, some `general' methods, that do not rely on task-specific operations, can be adapted to gaze following with appropriate modifications. For example, \RN{2} Model \cite{sajjadi2016regularization} enforces consistency on the outputs from the shared-weights teacher and student models, whose inputs are the same image added with different noises; Temporal Ensembling \cite{laine2017temporal} updates the teacher model output as the exponential moving average (EMA) of outputs from past epochs; Mean Teacher \cite{meanteacher} updates the teacher model weights as EMA of student model weights during training.  Although mostly tested on recognition tasks, the strategies of these methods can be adapted to gaze following and applied to the predicted heatmaps. Within these methods, the EMA-Teacher framework in Mean Teacher has been dominantly adopted by recent semi-supervised learning works spanning multiple tasks \cite{kwon2022semi, ot_crowdcounting, ho2024diffusion, assran2022masked, semi-vit}, despite usually augmented with task-specific operations for the specified tasks. Therefore, we choose Mean Teacher as the consistency regularization baseline in the experiments,
and we also show that we can use Mean Teacher to enhance the diffusion model training and achieve even larger improvements.

\section{Ablations in Grad-CAM Heatmap Computation} \label{ablation_gradcam}

In this section, we show the ablations of different alternatives for obtaining the Grad-CAM heatmaps from the VQA model, by evaluating the Grad-CAM heatmaps computed on the GazeFollow test set directly. We varied the decoder layer of the cross-attention map from which the Grad-CAM heatmaps are computed and tested overlaying the scene image with the head bounding box directly instead of the body bounding box obtained with a person detector. It can be seen from Tab.\ref{tab:gradcam_ablation} that even raw Grad-CAM heatmaps can achieve adequate performance on the GazeFollow test set. Overlaying the image with the body bounding box showed slightly better quality than overlaying with the head bounding box, while computing Grad-CAM from layer 11 (12 layers in total in the decoder of OFA model \cite{wang2022ofa}) showed the best performance.

\begin{table}[h]
\centering
\caption{\textbf{Quality of Grad-CAM heatmaps generated from different alternatives on GazeFollow test set.}}
\begin{tabular}{cc|ccc}
\Xhline{2\arrayrulewidth}
\multirow{2}{*}{Overlay} & \multirow{2}{*}{Decoder Layer}  & \multicolumn{2}{c}{Dist. $\downarrow$} & \multirow{2}{*}{AUC  $\uparrow$}                                  \\
                           &                         & Avg.                      & Min         &                            \\
\Xhline{2\arrayrulewidth}
Head Box & Layer 11 & 0.258 & 0.182 & 0.784  \\
Body Box & Layer 10 & 0.262 & 0.190 & 0.775 \\
Body Box & Layer 11 & \textbf{0.254} & \textbf{0.180} & \textbf{0.791} \\
Body Box & Layer 12 & 0.283 & 0.209 & 0.785 \\
\Xhline{2\arrayrulewidth}
\end{tabular}
\label{tab:gradcam_ablation}
\end{table}

\end{document}